\documentclass{article}
\usepackage{microtype}
\usepackage{graphicx}
\usepackage{booktabs}
\usepackage{amsmath,amssymb}
\usepackage{array}
\usepackage{tabularx}
\usepackage{enumitem}
\usepackage{listings}
\usepackage[hidelinks]{hyperref}
\usepackage{mlsys2025}
\ClearShipoutPicture
\usepackage{xcolor}
\usepackage[skins]{tcolorbox}
\usepackage{tikz}
\usepackage{pgfplots}
\pgfplotsset{compat=1.18}
\usepackage{adjustbox}
\usetikzlibrary{arrows.meta,fit,backgrounds,positioning,calc}
\usetikzlibrary{arrows.meta,positioning,fit,backgrounds,calc,decorations.pathreplacing,shapes.geometric,patterns}
\definecolor{mdl}{HTML}{E4DDF4}      \definecolor{mdlD}{HTML}{7B68B5}   % model-specific
\definecolor{hw}{HTML}{FCE4CC}       \definecolor{hwD}{HTML}{D9822B}    % hardware-specific
\definecolor{gen}{HTML}{DCEFD6}      \definecolor{genD}{HTML}{5E9E4A}   % generic mechanism
\definecolor{dat}{HTML}{DCEAF7}      \definecolor{datD}{HTML}{3A78B5}   % data / buffers
\definecolor{stt}{HTML}{FFF0C2}      \definecolor{sttD}{HTML}{B8860B}   % state
\definecolor{swp}{HTML}{D6F0EC}      \definecolor{swpD}{HTML}{23877A}   % swap window
\definecolor{q8}{HTML}{F8D7DA}       \definecolor{q8D}{HTML}{C0392B}    % FP8 tensor
\definecolor{f32}{HTML}{EDEDED}      \definecolor{f32D}{HTML}{6B6B6B}   % FP32 partials
\definecolor{ink}{HTML}{333333}
\tikzset{
  every picture/.style={font=\sffamily\small, line cap=round, line join=round},
  box/.style={draw=#1D, fill=#1, rounded corners=2.5pt, line width=.55pt,
              align=center, inner sep=3pt, font=\sffamily\footnotesize,
              execute at begin node={\hyphenpenalty=10000\exhyphenpenalty=10000}},
  box/.default=gen,
  tag/.style={font=\sffamily\scriptsize, text=black!60, inner sep=1pt},
  lbl/.style={font=\sffamily\footnotesize, inner sep=1pt},
  ttl/.style={font=\sffamily\bfseries\small, inner sep=1pt},
  flow/.style={-{Stealth[length=1.6mm]}, line width=.6pt, draw=black!60},
  dflow/.style={flow, dashed},
  thick flow/.style={-{Stealth[length=2mm]}, line width=1pt, draw=black!65},
  band/.style={draw=none, fill=#1, rounded corners=3pt},
  swapbox/.style={box=swp, dashed},
}

\usetikzlibrary{calc,fit,backgrounds}
\definecolor{rtblue}{HTML}{285F8F}
\definecolor{rtteal}{HTML}{23877A}
\definecolor{rtgold}{HTML}{C88824}
\definecolor{rtpale}{HTML}{EEF4F8}
\definecolor{rttakeawaybg}{HTML}{EDF7EF}
\definecolor{rttakeawayborder}{HTML}{B6D9BE}
\definecolor{rttakeawayaccent}{HTML}{387B52}
\newtcolorbox{takeaway}[1]{
  enhanced,
  colback=rttakeawaybg, colframe=rttakeawayborder,
  borderline west={1.5pt}{0pt}{rttakeawayaccent},
  boxrule=0.4pt, arc=1.5mm,
  boxsep=0pt, left=6pt, right=6pt, top=5pt, bottom=5pt,
  before skip=7pt plus 1pt, after skip=7pt plus 1pt,
  fontupper=\small,
  before upper={\textbf{Takeaway: #1}\par\smallskip}}
\newcommand{\code}[1]{\texttt{\small #1}}
\newcommand{\flashrt}{\textsc{FlashRT}}
\lstdefinestyle{capsulepseudo}{
  language=Python,
  basicstyle=\ttfamily\scriptsize\color{black!85},
  backgroundcolor=\color{rtpale},
  rulecolor=\color{rtblue!35},
  keywordstyle=\color{rtblue}\bfseries,
  emphstyle=\color{rtteal}\bfseries,
  stringstyle=\color{rtgold},
  commentstyle=\color{black!55}\itshape,
  emph={snapshot,replay,overwrite,restore},
  frame=single, framesep=4pt, xleftmargin=0pt, xrightmargin=0pt,
  breaklines=false, keepspaces=true, columns=fullflexible,
  aboveskip=4pt, belowskip=4pt, showstringspaces=false}

\newif\ifshowchecks
\showchecksfalse

\makeatletter\g@addto@macro{\UrlBreaks}{\do\-}\makeatother

\mlsystitlerunning{FlashRT: Fast Kernels Are Not Fast Pipelines}
\AtBeginDocument{\hypersetup{pdfauthor={Liang Su, David Chen, Jason Li, Kevin Guo, Teresa Lv}}}

\begin{document}
\twocolumn[
\mlsystitle{FlashRT: Fast Kernels Are Not Fast Pipelines\texorpdfstring{\\}{ }for Physical AI}
% \begin{center}\begin{minipage}{0.95\textwidth}{\small\color{red}\textbf{Kaiwen:} Use FlashRT as the system name; the subtitle already includes Physical AI, so the PAI suffix is redundant.\par}\end{minipage}\end{center}
\begin{center}
Liang Su\quad David Chen\quad Jason Li\quad Kevin Guo\quad Teresa Lv\par\vspace{4pt}
{\small AnyDevice AI \& Alps Lab}
\end{center}
\mlsyskeywords{Inference systems, physical AI, vision-language-action models, world action models, CUDA Graphs, quantization, GPUs}
\vskip 0.25in
\begin{abstract}
Vision-language-action (VLA), video-action (VA), and world action models (WAMs) form the backbone of modern physical AI systems, yet their real-time single-batch inference remains a defining bottleneck. Latency is dominated not by individual kernel computation, but by system-level overhead: thousands of short kernel launches, intermediate data movement across operator boundaries, and redundant recomputation across consecutive decisions. General-purpose graph compilers cannot address these costs, as they optimize individual calls in isolation and lack visibility into cross-call state semantics and global pipeline structure.
We present FlashRT, a low-latency inference engine built around \textbf{model-hardware pipeline co-design} rather than isolated kernel optimization. FlashRT combines three core mechanisms: deep kernel fusion with launch overlap eliminates inter-operator boundaries and launch gaps; an execution contract formalizes the pipeline’s stage DAG, update rates, and named buffers to enable cross-decision state reuse and selective refresh; and execution-state capsules enable fast snapshot, restore, and fork of persistent pipeline state. The system also supports drop-in acceleration of unmodified PyTorch models via parity- and latency-gated kernel replacement.
Evaluation covers six policies on RTX 5090 and four on Jetson AGX Thor. At matched 16-bit precision, FlashRT is \textbf{1.3--3.7$\times$ faster than official compiled paths}. Combining pipeline co-design with reduced precision delivers \textbf{3.38--12.66$\times$ speedup over official eager implementations}. In LIBERO-10, four-bit $\pi_{0.5}$ achieves 94.04\% success on Thor and 93.32\% on RTX 5090, compared with 92.4\% for the measured official FP16 reference. FlashRT is open source, distributed via Hugging Face Kernel Hub, and its opt-in $\pi_{0.5}$ kernels have been adopted into the LeRobot framework.
\end{abstract}

]
% Retain the review notice without an anonymous affiliation footnote.
\makeatletter
{\let\thefootnote\relax\footnotetext{\Notice@String}}
\makeatother

% ===========================================================================
\section{Introduction}
\label{sec:intro}
% \subsection{The Real-Time Deployment Gap for Physical AI}
Physical AI places learned policies in robot control loops: VLAs map images
and instructions to actions~\cite{pi05_2025,groot2025}, VAs generate video and
actions autoregressively~\cite{lingbotva2026}, and WAMs jointly denoise future
video and actions~\cite{flux3action2026,cosmos3_2026,motus2025}. Each call
returns an action chunk; its inference budget depends on the host's observation
and action-update schedule. We study this inference component on edge and
workstation GPUs, with task-level checks in LIBERO simulation.

Growing policies such as $\pi_{0.7}$ and FLUX 3 Action have 5--7B
parameters~\cite{pi07_2026,flux3action2026}. On Thor, official $\pi_{0.5}$
takes 222.64\,ms eager and 129.17\,ms compiled; FLUX 3 Action takes 1.35\,s
on RTX 5090 (Table~\ref{tab:e2e}). Reducing latency must also preserve action
fidelity, since numerical drift can change robot behavior.

% Physical AI places learned policies inside the control loop of robots. Three kinds
% of policy are in use. Vision-language-action (VLA) models map camera images, a
% language instruction, and robot state to future actions~\cite{pi05_2025,groot2025}.
% Video-action models (VAs) generate future video and actions chunk by chunk,
% autoregressively~\cite{lingbotva2026}. World action models (WAMs) denoise future
% video and actions together~\cite{flux3action2026,cosmos3_2026,motus2025}. In all
% three the next action chunk must be ready within the control period. Every decision
% runs at batch size one, on a module mounted on the robot, such as a Jetson AGX Thor,
% or on a workstation GPU beside it. Policies are also growing: $\pi_{0.7}$ has about
% 5B parameters and FLUX 3 Action 7B~\cite{pi07_2026,flux3action2026}. A faster implementation must return the same
% actions, since small numerical drift can change what the robot does.

Closing this gap takes more than faster individual kernels. We present
\flashrt{}, an inference engine for low-latency physical AI, guided by a simple
principle: \emph{fast kernels are necessary but not sufficient}~\cite{dehghani2021efficiency}. For each
model--hardware pipeline, \flashrt{} decides what can be fused,
overlapped, or reused across decisions; an execution contract makes those
choices executable. Four limitations motivate these choices.

First, \textbf{small kernels repeatedly materialize intermediates; deep kernel
fusion removes selected boundaries}. At batch size one, separate normalization,
residual, activation, and quantization kernels between general matrix
multiplications (GEMMs) can write values that the next kernel immediately
reads. TensorRT~\cite{tensorrt},
\code{torch.compile}~\cite{pytorch2}, and TVM~\cite{tvm2018} already fuse common
patterns, but some boundaries depend on deployment-specific facts, such as a
validated quantization scale and the next GEMM's input layout. \flashrt{} uses
these facts to hand-write the producer--consumer path, folding quantization into
the producer where precision and hardware permit. In $\pi_{0.5}$ on Thor, its
FP8 decoder layer uses 12 kernels versus 20 for \code{torch.compile}+FP8
(\S\ref{sec:fusion}).

Second, \textbf{remaining kernel dependencies need not serialize all work;
launch overlap hides part of their cost}. A successor can load weights before
it needs its predecessor's output, but an ordinary graph dependency waits for
the predecessor to finish. At eligible boundaries, \flashrt{} marks where the
successor may start and where it must wait, then applies CUDA's Programmatic
Dependent Launch (PDL)~\cite{cuda_launch_overlap}.
PDL is CUDA's mechanism; \flashrt{} supplies the safe kernel points and pipeline
order. On top of the already fused $\pi_{0.5}$ NVFP4 path, it further shortens
each decoder layer on Thor (\S\ref{sec:overlap}).

Third, \textbf{values needed by later decisions are often recomputed; frozen
execution state keeps them valid and reusable}. In the official $\pi_{0.5}$
implementation, step-dependent conditioning is rebuilt on every call;
LingBot-VA similarly recomputes cross-attention keys and values derived from an
unchanged instruction. CUDA Graph replay gives stable addresses but not the
application's rules for when such values become stale. \flashrt{} names its
fixed and persistent buffers, and records each
one's owner and refresh rule in the execution contract. It can replay captured
stages without per-kernel host dispatch and reuse prompt- or step-dependent
values across decisions. These named buffers also form an
\emph{execution-state capsule} that can be copied to snapshot, restore, or fork
a session (\S\ref{sec:capsule}).

Fourth, \textbf{optimized kernels are hard to adopt; checked replacements let an
existing host use them selectively}. Fast TensorRT deployments for policies
such as $\pi_{0.5}$ and GR00T require model-specific
integration~\cite{jetsonopenpi,grootdeploy}, and an existing PyTorch host needs a way to
check a replacement before using it. \flashrt{} packages each replaceable region
with a PyTorch reference and its kernels; the host keeps its own path unless the
replacement matches the reference and lowers end-to-end latency
(\S\ref{sec:distribute}). \flashrt{} kernels are distributed through the
Hugging Face Kernel Hub, and LeRobot adopted an opt-in $\pi_{0.5}$ kernel
integration~\cite{lerobotpr3870}.

We evaluate \flashrt{} on a Jetson AGX Thor for robot-side inference and an
RTX 5090 for workstation inference. Latency and numerical fidelity use fixed
or recorded inputs offline; closed-loop task success is evaluated for
$\pi_{0.5}$ and LingBot-VA in LIBERO simulation. We compare official implementations,
\code{torch.compile}, and available TensorRT deployments (\S\ref{sec:eval}),
and report public adoption (\S\ref{sec:impact}).

\subsection{Contributions}
% \flashrt{} is built, deployed, and measured. 
This paper makes three contributions:

\begin{itemize}
\item \textbf{An inference engine built around model--hardware pipeline co-design.}
We present \flashrt{}, an inference engine that bridges policy models and
production deployment through joint choices of GEMM tiling, precision,
producer--consumer layouts, fusion, and launch overlap. Rather than stopping
at isolated operator tuning, the design also accounts for stage boundaries,
input-update cadence, and state reuse. Its kernels support checked
integration into existing PyTorch hosts while retaining their control loop
and fallback path (\S\ref{sec:design}, \S\ref{sec:distribute}).

\item \textbf{Three complementary mechanisms for small-batch inference.}
Deep kernel fusion removes selected inter-operator memory round-trips, and launch
overlap reduces waiting at the boundaries that remain. The execution
contract records stage dependencies, input-update methods, and named
buffers for cross-decision reuse and selective refresh. Execution-state
capsules snapshot, restore, and fork declared continuation state through
device-local memory copies. In controlled
$\pi_{0.5}$/Thor runs, reuse reduces decision latency from 37.05 to
21.61\,ms; restoring context and generating the next action takes 9.73\,ms
instead of 21.48\,ms for rebuilding it. Reused and restored actions are
bit-exact (\S\ref{sec:exec}, \S\ref{sec:eval-hotpath}).

\item \textbf{Multi-policy evaluation and open-source impact.}
Across six policies on RTX 5090 and four on Thor, evaluated paths are
3.38--12.66$\times$ faster than official eager implementations and
1.3--3.7$\times$ faster than official compiled paths at matched 16-bit
precision. Paired LIBERO task checks cover $\pi_{0.5}$ and LingBot-VA.
\flashrt{} is open source, with kernels distributed through the Hugging Face
Kernel Hub, opt-in $\pi_{0.5}$ integration adopted in LeRobot, and public
developer and vendor acknowledgments
(\S\ref{sec:eval}--\S\ref{sec:impact}).

\end{itemize}

% ===========================================================================
\section{Background and Motivation}
\label{sec:background}
% \par{\small\color{red}\textbf{Kaiwen:} Strengthen the motivation by linking workload properties to bottlenecks that remain after compilation. Explain how fixed shapes, precision/layout constraints, and state update rates expose gaps in existing execution paths and motivate FlashRT's design; the current observations mainly enumerate optimization opportunities.\par}

\subsection{Three Classes of Policy Calls}
\label{sec:workload}
\begin{figure*}[!t]
\centering

\begin{adjustbox}{max width=\textwidth}

\begin{tikzpicture}[
  font=\small,
  lane/.style={
    font=\small\bfseries,
    anchor=east,
    align=right
  },
  box/.style={
    rounded corners=2pt,
    minimum height=9mm,
    align=center,
    inner sep=2.5pt,
    anchor=west
  },
  stage/.style={
    box,
    draw=rtblue!80,
    fill=rtpale
  },
  state/.style={
    box,
    draw=rtgold!85!black,
    fill=rtgold!12
  },
  swap/.style={
    box,
    draw=rtteal!85!black,
    fill=rtteal!10,
    dashed
  },
  arr/.style={
    -{Stealth[length=2mm]},
    line width=.7pt,
    draw=black!60
  },
  sarr/.style={
    -{Stealth[length=2mm]},
    line width=.8pt,
    draw=rtgold!85!black
  },
  capl/.style={
    font=\scriptsize,
    text=black!60
  },
  ptitle/.style={
    font=\small\bfseries,
    anchor=west
  },
  c0/.style={text width=14mm},
  c1/.style={text width=34mm},
  c2/.style={text width=25mm}
]

% ============================================================
% Row labels
% ============================================================

\node[lane] at (-0.2,0)
  {per\\prompt};

\node[lane] at (-0.2,-1.55)
  {per\\observation};

\node[lane] at (-0.2,-3.45)
  {per\\decision};

% ============================================================
% (a) VLA
% ============================================================

\begin{scope}[xshift=0cm]

\node[ptitle] at (0,0.9)
  {(a) VLA: $\pi_{0.5}$};

% Per prompt
\node[stage,c1] (a-p1) at (1.9,0)
  {instruction, time and\\style tables (precomputed)};

\node[state,c2] (a-p2) at (5.85,0)
  {static buffers};

% Per frame
\node[swap,c0] (a-f0) at (0,-1.55)
  {images\\(swap)};

\node[stage,c1] (a-f1) at (1.9,-1.55)
  {SigLIP (27 layers)\\+ prefix LM (18 layers)};

\node[state,c2] (a-f2) at (5.85,-1.55)
  {prefix K/V\\(state)};

% Per decision
\node[swap,c0] (a-d0) at (0,-3.45)
  {noise\\(swap)};

\node[stage,c1] (a-d1) at (1.9,-3.45)
  {action expert, 10 tokens\\$\times 10$ steps (small, many)};

\node[swap,c2] (a-d2) at (5.85,-3.45)
  {actions\\(readback)};

% Horizontal flow
\draw[arr] (a-p1) -- (a-p2);

\draw[arr] (a-f0) -- (a-f1);
\draw[arr] (a-f1) -- (a-f2);

\draw[arr] (a-d0) -- (a-d1);
\draw[arr] (a-d1) -- (a-d2);

% State feeds
\draw[sarr]
  (a-f2.south)
  -- ++(0,-0.3)
  -| ([xshift=-7mm]a-d1.north);

\draw[sarr]
  (a-p2.east)
  -- ++(0.4,0)
  |- ([xshift=7mm,yshift=3.5mm]a-d1.north)
  -- ([xshift=7mm]a-d1.north);

% Captured graph
\begin{scope}[on background layer]
\node[
  draw=black!45,
  dashed,
  rounded corners=3pt,
  fit=(a-d1),
  inner sep=2.5pt
] (a-cap) {};
\end{scope}

\node[capl,anchor=north west]
  at (a-cap.south west)
  {captured graph, replayed per decision};

\end{scope}

% ============================================================
% (b) VA
% ============================================================

\begin{scope}[xshift=9.5cm]

\node[ptitle] at (0,0.9)
  {(b) VA: LingBot-VA};

% Per prompt
\node[stage,c1] (b-p1) at (1.9,0)
  {text encoder\\(once per prompt)};

\node[state,c2] (b-p2) at (5.85,0)
  {text condition\\(static)};

% Per frame
\node[swap,c0] (b-f0) at (0,-1.55)
  {frames\\(swap)};

\node[stage,c1] (b-f1) at (1.9,-1.55)
  {VAE encode\\observed frames};

\node[state,c2] (b-f2) at (5.85,-1.55)
  {past-chunk K/V\\(state, grows)};

% Per decision
\node[swap,c0] (b-d0) at (0,-3.45)
  {noise\\(swap)};

\node[stage,c1] (b-d1) at (1.9,-3.45)
  {video + action tokens\\$\times N$ steps (heavy, few)};

\node[swap,c2] (b-d2) at (5.85,-3.45)
  {actions\\(readback)};

% Horizontal flow
\draw[arr] (b-p1) -- (b-p2);

\draw[arr] (b-f0) -- (b-f1);
\draw[arr] (b-f1) -- (b-f2);

\draw[arr] (b-d0) -- (b-d1);
\draw[arr] (b-d1) -- (b-d2);

% State feeds
\draw[sarr]
  (b-f2.south)
  -- ++(0,-0.3)
  -| ([xshift=-7mm]b-d1.north);

\draw[sarr]
  (b-p2.east)
  -- ++(0.4,0)
  |- ([xshift=7mm,yshift=3.5mm]b-d1.north)
  -- ([xshift=7mm]b-d1.north);

% Captured graph
\begin{scope}[on background layer]
\node[
  draw=black!45,
  dashed,
  rounded corners=3pt,
  fit=(b-d1),
  inner sep=2.5pt
] (b-cap) {};
\end{scope}

\node[capl,anchor=north west]
  at (b-cap.south west)
  {captured graph, replayed per decision};

\end{scope}

% ============================================================
% (c) WAM
% ============================================================

\begin{scope}[xshift=19cm]

\node[ptitle] at (0,0.9)
  {(c) WAM: FLUX 3 Action};

% Per prompt
\node[stage,c1] (c-p1) at (1.9,0)
  {Qwen3-VL-4B\\(once per prompt)};

\node[state,c2] (c-p2) at (5.85,0)
  {text condition\\(static)};

% Per frame
\node[swap,c0] (c-f0) at (0,-1.55)
  {views\\(swap)};

\node[stage,c1] (c-f1) at (1.9,-1.55)
  {video VAE\\encode};

\node[state,c2] (c-f2) at (5.85,-1.55)
  {frame latents\\(static)};

% Per decision
\node[swap,c0] (c-d0) at (0,-3.45)
  {noise\\(swap)};

\node[stage,c1] (c-d1) at (1.9,-3.45)
  {7B DiT, 32 actions\\$\times N$ steps (heavy, few)};

\node[swap,c2] (c-d2) at (5.85,-3.45)
  {actions\\(readback)};

% Horizontal flow
\draw[arr] (c-p1) -- (c-p2);

\draw[arr] (c-f0) -- (c-f1);
\draw[arr] (c-f1) -- (c-f2);

\draw[arr] (c-d0) -- (c-d1);
\draw[arr] (c-d1) -- (c-d2);

% State feeds
\draw[sarr]
  (c-f2.south)
  -- ++(0,-0.3)
  -| ([xshift=-7mm]c-d1.north);

\draw[sarr]
  (c-p2.east)
  -- ++(0.4,0)
  |- ([xshift=7mm,yshift=3.5mm]c-d1.north)
  -- ([xshift=7mm]c-d1.north);

% Captured graph
\begin{scope}[on background layer]
\node[
  draw=black!45,
  dashed,
  rounded corners=3pt,
  fit=(c-d1),
  inner sep=2.5pt
] (c-cap) {};
\end{scope}

\node[capl,anchor=north west]
  at (c-cap.south west)
  {captured graph, replayed per decision};

\end{scope}

\end{tikzpicture}

\end{adjustbox}

\caption{One decision of (a) a VLA, (b) a VA, and (c) a WAM, with work grouped by how
often it has to run. Results of the upper rows are kept in static buffers and state
regions that the per-decision stage only reads; inputs and outputs pass through swap
windows, buffers whose contents are replaced without touching the captured graphs.
The rows are the same for all three kinds of policy; what differs is the cost of a
denoising step and the state: the VA keeps a key/value cache over earlier chunks that
grows across decisions. A monolithic graph replayed unchanged reruns the per-prompt row at every
decision. Dashed outlines mark captured CUDA Graphs. \flashrt{} records this
structure in the execution contract (\S\ref{sec:contract}).}

\label{fig:tick}

\end{figure*}
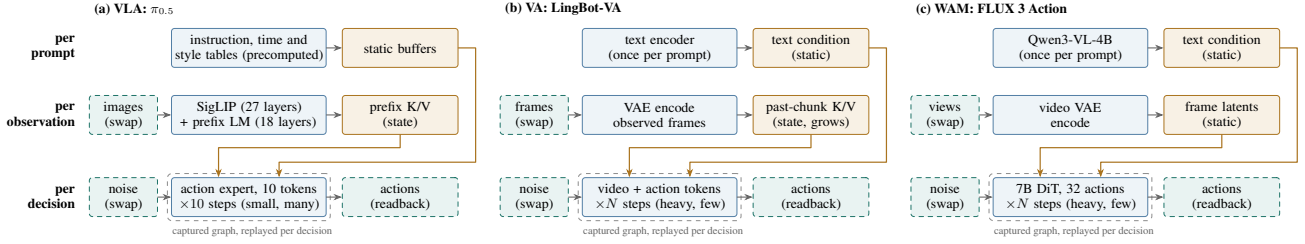

%Figure~\ref{fig:tick} shows one decision of each kind.
Figure~\ref{fig:tick} illustrates one decision cycle for each policy class. In $\pi_{0.5}$ (Figure~\ref{fig:tick}a),
a 27-layer SigLIP encoder embeds each camera view, an 18-layer PaliGemma language model
reads the image and prompt tokens (a prefix of about 500 to 800 tokens) and writes its
key/value cache, and an 18-layer action expert attends to that cache while integrating
ten flow-matching steps over a chunk of ten action tokens~\cite{pi0_2024,pi05_2025,
flowmatching2022}. GR00T N1.7 has the same overall structure~\cite{groot2025}.
A VA generates video and actions autoregressively, one chunk at a time. In LingBot-VA
(Figure~\ref{fig:tick}b), a text encoder reads the instruction once per prompt, a VAE
encodes the observed frames, and the backbone denoises the video and action tokens of
the next chunk while attending to a key/value cache over earlier chunks; the cache
persists from one decision to the next and grows up to a fixed window~\cite{lingbotva2026}.
A WAM also starts from a video generation model. In FLUX 3 Action
(Figure~\ref{fig:tick}c), a frozen Qwen3-VL-4B encodes the instruction, a frozen video
VAE encodes the camera frames, and a 7B diffusion transformer denoises future frames
together with a chunk of 32 actions~\cite{flux3action2026}. Cosmos and
Motus follow the same pattern with other backbones~\cite{cosmos3_2026,motus2025}.

All the policy calls share three structural properties:
First, parts of the call change at different rates: the instruction once per task, the observation once per decision, and the denoising loop several times within each decision.
Second, a deployment fixes the number of views, context-length bucket, action
horizon, and denoising-step count. These constraints allow fixed-shape stages
to be prepared once and replayed.
Third, shapes vary widely within one call. The $\pi_{0.5}$ action expert
processes ten rows per step, whereas VA and WAM backbones process thousands
of video and action tokens. These workloads need different GEMM tiling and
fusion choices; a VA also carries history state across decisions.
% \end{enumerate}
Fixed shapes make replay possible, but they do not by themselves determine
the fastest execution path. For these policies, action shapes, the formats
exchanged between adjacent kernels, and the lifetime of prepared values
guide the design choices below.

% All three kinds of call share three properties. First, parts of the call change at
% different rates: the instruction once per task, the images once per camera frame, and
% only the denoising loop at every decision. Second, everything else is fixed for a given
% deployment, including the number of views, the prefix or context length, the action
% horizon, and the number of denoising steps, so the whole call can be prepared once and
% then replayed. Third, shapes vary widely within one call. Where the three kinds differ is the cost of a denoising step: the action expert of a VLA processes ten rows per step, whereas a VA or WAM step processes thousands of video and action tokens. A VLA is dominated by many small steps, a VA or WAM by fewer, heavier ones, and a VA also carries state that grows across decisions.

% We report per-decision latency, the numerical agreement of the returned actions with a
% reference, and task success. All robot results in this paper are offline, on recorded
% observations and in LIBERO simulation~\cite{libero2023}; we did not run the policies on
% a physical robot.

\subsection{Kernel Boundaries and Cross-Decision Reuse}
\label{sec:motivation}
More capable models alone do not solve physical AI deployment. On both
edge devices and workstation GPUs, policy calls remain small-batch and
latency-sensitive.
Three costs recur at batch size one: intermediate traffic between kernels,
waiting at the boundaries that remain, and recomputation of unchanged values
across calls. Table~\ref{tab:breakdown} profiles intermediate-kernel
costs in one complete $\pi_{0.5}$ decision of the official OpenPI PyTorch
eager implementation on Jetson AGX Thor; Figure~\ref{fig:deep-fusion} separately tests launch overlap.
Within-call timing cannot show whether a
value needed by the next decision was retained or recomputed.
% LLV's incoming deployment-motivation draft, integrated above:
% \llv{The biggest bottleneck in physical AI right now isn’t model intelligence — it’s deployment: all facing small-batch, latency-sensitive problems, not matter on GPU or edge. }

\begin{table}[t]
\centering
\caption{Estimated time breakdown of one $\pi_{0.5}$ decision with the official OpenPI eager
implementation on Jetson AGX Thor. Kernels between GEMMs are the small kernels that
normalize, add residuals, change layouts, or convert precision; gaps are the time when no
kernel runs. Because profiling inflates the gaps, we estimate them as the unprofiled total
latency minus the measured kernel time.}
\label{tab:breakdown}
\footnotesize
\setlength{\tabcolsep}{4pt}
\begin{tabular}{@{}lrr@{}}
\toprule
 & ms & \% \\
\midrule
Kernel execution & 172.50 & 77.48 \\
\quad of which, kernels between GEMMs & 70.10 & 31.49 \\
Gaps between kernels (estimated) & 50.14 & 22.52 \\
\midrule
Total & 222.64 & 100 \\
\bottomrule
\end{tabular}
\end{table}

\textbf{Observation 1: materialized intermediates cost time and traffic.}
The kernel trace records 70.10\,ms in non-matrix kernels between GEMMs,
about 31.49\% of the unprofiled decision latency in this estimate. They normalize, add a residual, transform a layout, or convert precision.
Each separate kernel must start and move its intermediate through memory.
Folding eligible operations into neighboring kernels reduces launches and intermediate memory traffic.
Figure~\ref{fig:deep-fusion} shows that the problem remains after compilation: with
\code{torch.compile} and FP8, one action-expert layer still runs 20 kernels. Whether a boundary can be removed depends on the
producer's output and its consumer's precision and layout
(\S\ref{sec:fusion}).

\textbf{Observation 2: a remaining dependency need not serialize all work.}
Estimated gaps account for 22.52\% of an eager decision (Table~\ref{tab:breakdown}),
and fewer kernels leave fewer gaps (\S\ref{sec:fusion}). Some boundaries must
remain, however. At each of them, a plain dependency makes the next kernel wait
for its predecessor to end before it starts any work, even setup or weight
loading that does not need the predecessor's output. Figure~\ref{fig:deep-fusion} bottom shows that such independent work can start early
while the dependency is still respected (\S\ref{sec:overlap}).

\textbf{Observation 3: results needed later can be lost at a call boundary.}
Figure~\ref{fig:tick} shows that an unchanged instruction can supply many
decisions. Yet a
call-scoped path may recompute the embeddings, tables, or K/V derived from it on every call,
or recycle an intermediate buffer as soon as its last consumer in that call
finishes. In the official $\pi_{0.5}$ implementation, for example, the time
conditioning of each denoising step depends only on the step index, yet it is
recomputed at every step of every call. Keeping such values cuts a decision from
37.05 to 21.61\,ms (\S\ref{sec:eval-hotpath}). Discarding such values repeats work; retaining
them without a refresh rule risks using stale data after the instruction or
frame changes. The missing information is how long each value remains valid
and who may overwrite it (\S\ref{sec:exec}).

These opportunities require scale, shape, and layout for fusion; safe trigger
and wait points for overlap; and update rates, validity rules, and owners for
reuse (\S\ref{sec:fusion}--\S\ref{sec:exec}).

% ===========================================================================
\section{Design of \flashrt{}}
\label{sec:design}
% \par{\small\color{red}\textbf{Kaiwen:} Bring the key co-design tradeoffs from Appendix~\ref{app:fusion} into this section. For example, the initial NVFP4 fusion increased downstream weight traffic, while a compact output layout made the pipeline faster. Explain how model-specific shapes and precision requirements drive GPU tiling, layout, and fusion choices, so the main text substantiates the model--hardware co-design contribution.\par}
\flashrt{} chooses stage boundaries, update rates, shapes, layouts, precision,
and capture granularity together. Hand-written kernels implement fusion and
launch overlap; an execution contract then records the completed stage DAG
and buffer rules for replay and capsules.
The \flashrt{} engine owns the captured graphs and fixed-address buffers.
A session manager uses the exported contract to select stage replays when the
application reports changed inputs, and copies declared state for capsules.

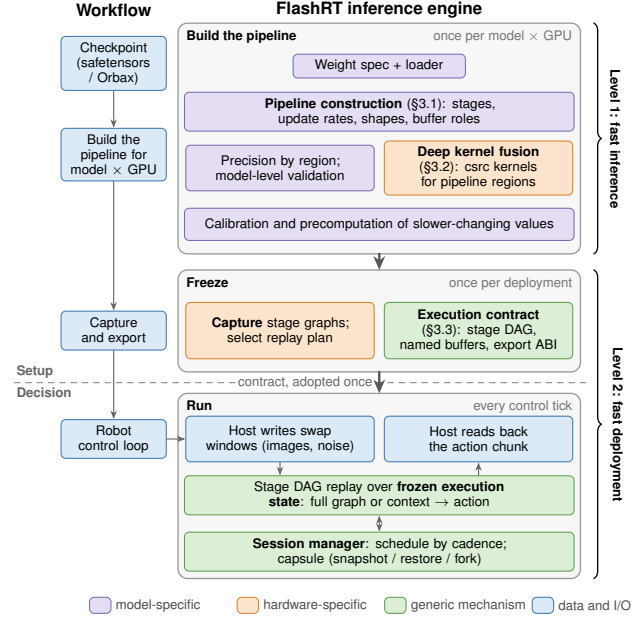
\begin{figure}[!t]
\centering
\resizebox{\columnwidth}{!}{% Fig. 1 (v5)  Direct revision of fig1_architecture_revised.tex in its own layout and palette:
% level labels in black, no ecosystem column, and the stack grouped into three modules
% (build, freeze, run) with the sub-modules inside them; arrows carry only the flow.
\begin{tikzpicture}[x=1cm,y=1cm,
  sbox/.style={box=#1, text width=3.3cm, minimum height=8.5mm, font=\sffamily\scriptsize},
  wbox/.style={box=#1, text width=7.05cm, minimum height=6mm, font=\sffamily\scriptsize},
  module/.style={draw=black!40, fill=black!3, rounded corners=5pt, line width=.7pt},
  mtitle/.style={font=\sffamily\bfseries\scriptsize, anchor=north west, inner sep=0pt},
  msub/.style={font=\sffamily\scriptsize, text=black!55, anchor=north east, inner sep=0pt},
  big/.style={-{Stealth[length=2.6mm, width=2.2mm]}, line width=1.3pt, draw=black!65},
  lvl/.style={font=\sffamily\bfseries\scriptsize, text=black}]
% Column headers and setup/decision boundary.
\node[ttl] at (1.85,11.1) {Workflow};
\node[ttl] at (6.8,11.1) {FlashRT inference engine};
\draw[black!45, dashed, line width=.6pt] (0.0,4.15) -- (10.75,4.15);
\node[tag, anchor=south west] at (0.0,4.2) {\textbf{Setup}};
\node[tag, anchor=north west] at (0.0,4.1) {\textbf{Decision}};

% Workflow, one box per module.
\node[box=dat, text width=1.75cm, font=\sffamily\scriptsize] (w1) at (1.85,10.1) {Checkpoint\\(safetensors / Orbax)};
\node[box=dat, text width=1.75cm, font=\sffamily\scriptsize] (w2) at (1.85,8.4) {Build the pipeline for model $\times$ GPU};
\node[box=dat, text width=1.75cm, font=\sffamily\scriptsize] (w3) at (1.85,5.12) {Capture and export};
\node[box=dat, text width=1.75cm, font=\sffamily\scriptsize] (w4) at (1.85,3.1) {Robot control loop};
\draw[flow] (w1) -- (w2); \draw[flow] (w2) -- (w3); \draw[flow] (w3) -- (w4);

% ---------------- module 1: build the pipeline ----------------------------
\draw[module] (3.05,6.55) rectangle (10.55,10.85);
\node[mtitle] at (3.2,10.72) {Build the pipeline};
\node[msub] at (10.4,10.72) {once per model $\times$ GPU};
\node[box=mdl, text width=3.0cm, font=\sffamily\scriptsize] (ld) at (6.8,10.05) {Weight spec + loader};
\node[wbox=mdl] (pd) at (6.8,9.2) {\textbf{Pipeline construction} (\S3.1): stages, update rates, shapes, buffer roles};
\node[sbox=mdl] (pr) at (4.95,8.15) {Precision by region; model-level validation};
\node[sbox=hw] (kp) at (8.65,8.15) {\textbf{Deep kernel fusion} (\S3.2): csrc kernels for pipeline regions};
\node[wbox=mdl] (cb) at (6.8,7.1) {Calibration and precomputation of slower-changing values};

% ---------------- module 2: freeze ----------------------------------------
\draw[module] (3.05,4.35) rectangle (10.55,6.25);
\node[mtitle] at (3.2,6.12) {Freeze};
\node[msub] at (10.4,6.12) {once per deployment};
\node[sbox=hw, minimum height=10.5mm] (cp) at (4.95,5.12) {\textbf{Capture} stage graphs; select replay plan};
\node[sbox=gen, minimum height=10.5mm] (ct) at (8.65,5.12) {\textbf{Execution contract} (\S3.3): stage DAG, named buffers, export ABI};

% ---------------- module 3: run every control tick ------------------------
\draw[module] (3.05,0.5) rectangle (10.55,3.95);
\node[mtitle] at (3.2,3.82) {Run};
\node[msub] at (10.4,3.82) {every control tick};
\node[sbox=dat] (hin) at (4.95,3.1) {Host writes swap windows (images, noise)};
\node[sbox=dat] (hout) at (8.65,3.1) {Host reads back the action chunk};
\node[wbox=gen] (dag) at (6.8,2.05) {Stage DAG replay over \textbf{frozen execution state}: full graph or context $\rightarrow$ action};
\node[wbox=gen] (session) at (6.8,1.0) {\textbf{Session manager}: schedule by cadence; capsule (snapshot / restore / fork)};
\draw[flow] (hin.south) -- (hin.south |- dag.north);
\draw[flow] (hout.south |- dag.north) -- (hout.south);
\draw[flow, {Stealth[length=1.6mm]}-{Stealth[length=1.6mm]}] (dag) -- (session);
\draw[flow] (w4.east) -- (hin.west);

% ---------------- flow between modules ------------------------------------
\draw[big] (6.8,6.55) -- (6.8,6.25);
\draw[big] (6.8,4.35) -- node[pos=0.5, left=2pt, tag, fill=white]{contract, adopted once} (6.8,3.95);

% ---------------- levels (black) and legend --------------------------------
\draw[decorate, decoration={brace, amplitude=4pt}, line width=.7pt, draw=black]
  (10.8,10.85) -- (10.8,6.55) node[midway, right=5pt, lvl, rotate=-90, anchor=south, fill=white]{Level 1: fast inference};
\draw[decorate, decoration={brace, amplitude=4pt}, line width=.7pt, draw=black]
  (10.8,6.25) -- (10.8,0.5) node[midway, right=5pt, lvl, rotate=-90, anchor=south, fill=white]{Level 2: fast deployment};
\foreach \c/\t [count=\i from 0] in {mdl/model-specific, hw/hardware-specific, gen/generic mechanism, dat/{data and I/O}}{
  \node[box=\c, minimum width=4mm, minimum height=3mm, inner sep=0pt] at (1.6+\i*2.75,0.0) {};
  \node[tag, anchor=west, text=black!70] at (1.85+\i*2.75,0.0) {\t};}
\end{tikzpicture}}
\vspace{-10pt}
\caption{Building and running a \flashrt{} model--hardware pipeline. Setup constructs
the pipeline from a checkpoint, chooses stage boundaries and precision, validates
the model, and captures the selected stages. The execution contract records the
stage DAG, named buffers, and export interface. At each control tick, the host
fills swap windows and reads the action chunk while the session manager schedules
replay over frozen execution state and can copy declared state for capsules.}
\label{fig:design}
\end{figure}

\subsection{Architecture and Model Integration}
\label{sec:pipeline}
Figure~\ref{fig:design} maps a loaded checkpoint into stages with declared
inputs, outputs, shapes, buffer roles, and update rates. Instruction-dependent
work runs at prompt setup; vision, prefix, and action work run per decision.
Setup selects kernels and precision, calibrates scales, checks fidelity,
prepares slower-changing values, and chooses launch boundaries. It captures
the full decision or split context/action stages, then exports the contract.

At decision time, the robot loop fills swap windows with observations and noise;
the session manager replays the selected captured stages over fixed-address
buffers; and the host reads the action chunk. Declared state regions can be
copied as capsules. Integration into an existing host is described in
\S\ref{sec:distribute}.

Across models, the reusable components are the buffer roles, contract and
kernel interfaces, and qualification procedure. Stage decomposition, precision,
layouts, and trigger/wait placement are designed and validated per model--hardware
pipeline; the contract records those choices after design.

\subsection{Deep Kernel Fusion}
\label{sec:fusion}
\label{sec:precision}
\begin{figure*}[!t]
\centering
\resizebox{\textwidth}{!}{\input{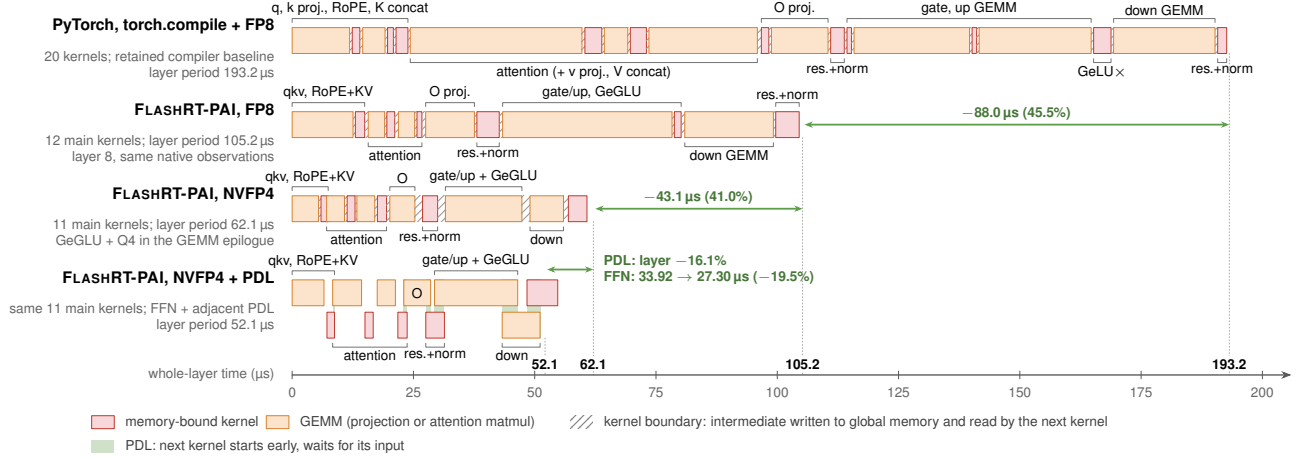}}
\vspace{-10pt}
\caption{Kernel timelines for one $\pi_{0.5}$ action-expert decoder layer on
Jetson AGX Thor (10 action rows), drawn on one time axis. The upper rows compare
the PyTorch FP8 baseline with \flashrt{}'s FP8 and NVFP4 pipelines. Bars show measured
CUPTI traces under CUDA Graph replay, and hatching marks kernel boundaries. The
bottom pair holds the 11-kernel NVFP4 path fixed and enables programmatic dependent
launch (PDL) at the FFN and adjacent O/QKV boundaries; green marks overlap.
Separately, the PyTorch step recomputes 37 AdaRMS modulation projections in a
1.87\,ms FP32-weight kernel; \flashrt{} prepares them once per prompt.}
\label{fig:deep-fusion}
\end{figure*}
\flashrt{} fuses memory-bound work around compute-heavy kernels: examples
include residual addition with normalization and output quantization, RoPE
with QKV split and cache write, and patch embedding with layout conversion.
Compute-heavy GEMMs and attention keep their specialized implementations. We
select a fused boundary when a supported fixed shape and layout let the producer
emit the consumer's required format, all consumers remain valid, and the
complete path passes fidelity and latency checks. Other boundaries remain
separate.

\textbf{Let the action shape determine the GEMM tile.}
The $\pi_{0.5}$ action expert has ten activation rows, while the stock NVFP4
GEMM tile reserves space for 128. For selected decoder projections,
\flashrt{} swaps the operand roles so that the large tile dimension carries
weights and the short dimension carries the ten action rows. The output
store also matches the next kernel's input
layout, avoiding a separate transpose. The deployment shape therefore
determines both the GEMM tile and the producer--consumer interface
(Appendix~\ref{app:fusion}).

\textbf{Choose fusion and the consumer's layout together.}
For an encoder FFN, \flashrt{} packs gate/up weights together so the GEMM
can apply the gated activation and produce packed four-bit values directly.
The first fused version wrote a full-width result and doubled the next
GEMM's weight traffic, making the feed-forward chain slower. A compact,
half-width output preserved the consumer's original shape and made the
chain faster. It also quantized the intermediate once rather than twice.
The useful optimization is the complete producer--consumer path: fewer
launches help only when the output format also suits the next kernel.
Appendix~\ref{app:fusion} gives the layout details.

Figure~\ref{fig:deep-fusion} traces a complete $\pi_{0.5}$ action-expert
decoder layer on Jetson AGX Thor. At FP8 GEMM precision, \flashrt{} reduces the
layer from 20 kernels in the PyTorch \code{torch.compile} path to 12. It fuses
the small operations around the FFN's two GEMMs: residual addition,
normalization, and input quantization share one kernel, while GeGLU and output
quantization share another (\S\ref{sec:eval-fusion}). The NVFP4 path also folds GeGLU into a GEMM
epilogue, leaving 11 kernels per layer.
At the fused quantization boundaries in Figure~\ref{fig:deep-fusion}, the
producer emits the GEMM input and its scale without a separate activation
read/write pass. This requires a supported precision, shape, and layout;
$\pi_{0.5}$ still has an attention-output path with separate quantization.
Selected low-bit paths use activation-aware weight scaling~\cite{awq2024}.
Scales are calibrated before capture, cached by checkpoint and sequence
length, and checked against model-level fidelity results
(\S\ref{sec:eval-fidelity}).

\subsection{Launch Overlap}
\label{sec:overlap}
Fusion leaves some producer--consumer boundaries intact. Overlap requires
input-independent successor work, a safe wait before dependent reads, and a
measured gain despite resource contention. At eligible pairs,
\flashrt{} uses CUDA's programmatic dependent launch (PDL) to start a successor
GEMM's weight loading before its activation is ready, then waits before
reading that activation. The pipeline selects the pair, and the kernels mark
safe trigger and wait points that an ordinary graph dependency does not
expose~\cite{cuda_launch_overlap}. PDL permits overlapping execution but does
not guarantee it; the successor still waits before it reads the
predecessor's output. In the bottom two rows of
Figure~\ref{fig:deep-fusion}, enabling PDL at the FFN and adjacent O/QKV
boundaries shortens the already fused 11-kernel NVFP4 layer from 62.1 to
52.1\,$\mu$s with the same kernels and precision. CUDA supplies PDL; our design
chooses and validates the model-specific trigger and wait points.
Appendix~\ref{app:overlap} gives three cases, including boundaries where more
overlap does not help.

\subsection{Execution Interface and Reusable State}
\label{sec:exec}
\label{sec:contract}
\label{sec:hotpath}
The execution contract exposes a completed pipeline through a shared execution
interface. Reuse is valid only while the inputs that produced a retained value
are unchanged. The interface records stage order, supported shapes, and named
buffers with owners, sizes, and update methods. The engine checks stage
references and selects a captured replay for the requested shape.
Appendix~\ref{app:contract} gives the execution rules, interface fields, and a
worked $\pi_{0.5}$ plan. \emph{Static}
buffers hold prepared values; \emph{swap windows} hold replaceable inputs and
outputs at fixed addresses; \emph{state regions} retain mutable state between
stages or decisions. The engine allocates these buffers once and captures
selected stages as CUDA Graphs keyed by shape.

The interface records stage order, input-update methods, and the buffer
regions needed for continuation. In $\pi_{0.5}$, instruction embeddings can
be reused while the instruction is unchanged, whereas a new observation
requires rebuilding image-conditioned K/V. If another session overwrites
the working buffers, restoring the saved regions lets the original session
continue without rebuilding its completed context.

\begin{figure}[t]
\centering
\resizebox{0.8\columnwidth}{!}{% Fixed-buffer execution within and across pi_0.5 decisions.
% Each decision consumes a fresh image. Only instruction embeddings and fixed
% timestep-conditioning tables are carried across calls in this schematic.
\begin{tikzpicture}[x=1cm,y=1cm,
  stepbox/.style={draw=black!40, fill=white, rounded corners=1.5pt,
                  font=\sffamily\scriptsize, align=left, anchor=west,
                  text width=3.45cm, inner sep=2pt, minimum height=7mm},
  buf/.style={draw=black!30, fill=#1, rounded corners=1.5pt,
              font=\sffamily\scriptsize, align=left, anchor=west,
              text width=3.45cm, inner sep=2pt, minimum height=7mm},
  hdr/.style={font=\sffamily\bfseries\scriptsize,
              text=black!75, anchor=west, inner sep=0pt},
  smalllabel/.style={font=\sffamily\scriptsize, text=black!65,
                     anchor=west, align=left, inner sep=0pt},
  legend/.style={font=\sffamily\tiny, text=black!65,
                 anchor=west, inner sep=0pt},
]
\def\stepnum#1{\textcircled{\tiny #1}}
\node[hdr] at (0,7.00) {(a) Inside decision $d$};
\node[hdr] at (0,6.57) {decision order};
\node[hdr] at (4.55,6.57) {fixed GPU buffers};
\node[stepbox] at (0,5.96) {\stepnum{1} Fill image and noise};
\node[stepbox] at (0,5.04) {\stepnum{2} Run context (image + prompt)};
\node[stepbox] at (0,4.12) {\stepnum{3} Run action stage (denoise)};
\node[stepbox] at (0,3.20) {\stepnum{4} Read action chunk};
\foreach \yy in {5.53,4.61,3.69}{
  \draw[flow] (1.90,\yy) -- (1.90,\yy-.16);}
\node[buf=swp] at (4.55,5.96) {images: \stepnum{1} fill; \stepnum{2} read};
\node[buf=swp] at (4.55,5.04) {noise $\rightarrow$ actions: \stepnum{1} fill; \stepnum{3} update; \stepnum{4} read};
\node[buf=stt] at (4.55,4.12) {encoder K/V: \stepnum{2} write; \stepnum{3} read across denoising steps};
\node[buf=dat, minimum height=10mm] (prepared) at (4.55,3.07)
  {prepared once:\\prompt embeddings: \stepnum{2} read\\step tables: \stepnum{3} read};

\node[draw=black!30, fill=swp, minimum width=2.6mm, minimum height=2.2mm,
      inner sep=0pt] at (0.12,2.35) {};
\node[legend] at (0.34,2.35) {refilled buffer};
\node[draw=black!30, fill=stt, minimum width=2.6mm, minimum height=2.2mm,
      inner sep=0pt] at (2.55,2.35) {};
\node[legend] at (2.77,2.35) {retained K/V};
\node[draw=black!30, fill=dat, minimum width=2.6mm, minimum height=2.2mm,
      inner sep=0pt] at (4.75,2.35) {};
\node[legend] at (4.97,2.35) {prepared once};
\draw[black!35, densely dashed, line width=.5pt] (0,1.99) -- (8.25,1.99);
\node[legend, fill=white, inner sep=2pt] at (4.55,1.99) {call boundary};

% The same stage sequence runs on a fresh image in the next call. The neutral
% connector carries the blue prepared values, never image-conditioned K/V.
\node[hdr] at (4.55,1.43) {(b) Across calls: decision $d+1$};
\node[smalllabel] at (4.55,1.05) {Same instruction; fresh image};
\node[stepbox] at (0,1.10) {\stepnum{1} Fill new image and noise};
\node[stepbox] (nextcontext) at (0,0.18) {\stepnum{2} Run context (image + prompt)};
\node[stepbox] (nextaction) at (0,-0.74) {\stepnum{3} Run action stage (denoise)};
\node[stepbox] at (0,-1.66) {\stepnum{4} Read next action chunk};
\draw[flow] (1.90,2.83) -- (1.90,1.49);
\foreach \yy in {0.67,-0.25,-1.17}{
  \draw[flow] (1.90,\yy) -- (1.90,\yy-.16);}
\draw[draw=black!55, line width=.6pt]
  (prepared.east) -- (8.55,3.07) -- (8.55,-0.74);
\draw[flow] (8.55,0.18) -- (nextcontext.east);
\draw[flow] (8.55,-0.74) -- (nextaction.east);
\node[smalllabel, fill=white, inner sep=2pt] at (4.55,0.18)
  {reuse prompt embeddings};
\node[smalllabel, fill=white, inner sep=2pt] at (4.55,-0.74)
  {reuse step tables};
\node[smalllabel] at (4.55,-1.66) {New image $\rightarrow$ new encoder K/V};
\end{tikzpicture}}
\caption{Fixed-buffer reuse in $\pi_{0.5}$. (a) Circled numbers link stages to
their buffer accesses; yellow K/V is shared by the denoising steps within a call.
(b) The next call processes a fresh image and rebuilds K/V, while reusing the
unchanged instruction's embeddings and fixed step-conditioning tables. The
connector carries these prepared values across the call boundary.}
\label{fig:contract-execution}
\end{figure}
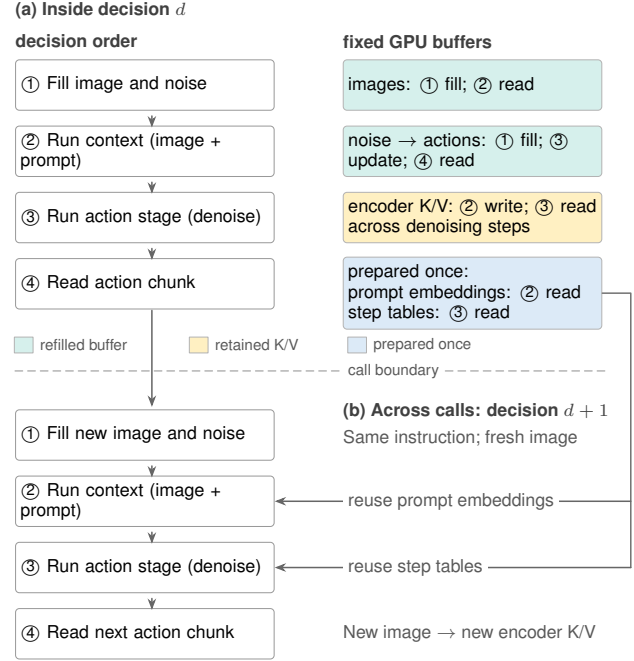

\noindent\textbf{Within a decision.}
Figure~\ref{fig:contract-execution}~(a) shows the split $\pi_{0.5}$ plan. The host
fills image and noise buffers; context encodes the image and prompt into K/V;
action reads that K/V throughout its denoising steps and transforms the noise
buffer into the action chunk. Shared named buffers let successive captured
stages consume results directly. The session manager orders stage replays
according to the contract; captured stages use fixed addresses without
replay-time allocation or host-dependent choices. The full-call plan and the
split plan return identical actions for identical inputs and noise.

\noindent\textbf{Across decisions.}
Figure~\ref{fig:contract-execution}~(b) carries prepared values into the next call.
The normal $\pi_{0.5}$ inference path uploads the supplied images and reruns
vision and context on every call; it reuses unchanged instruction embeddings
and fixed step-conditioning tables. Even the encoder's text-token K/V depends
on the image and is rebuilt. Prompt and robot-state updates enter through the
model's supported input methods. A host driving context and action separately
must refresh context for the current inputs before generating the next action.
At a stage boundary, \code{snapshot} copies declared restorable regions into
an \emph{execution-state capsule}; \code{restore} copies them back, and
\code{fork} copies them into another session.\label{sec:capsule}
The export lists region names, offsets, and lengths; loading checks the schema
and plan fingerprint, and resumption must respect the model's chunk boundaries.
For $\pi_{0.5}$, the capsule holds the context K/V, the prompt embedding, and the
decoder position table (14\,MB). For LingBot-VA, it also holds the history K/V,
the VAE cache, and the random-number state (1.4\,GB).
After another session overwrites the working buffers, restoring the capsule
reproduces the next actions bit for bit, without recapture and without
rebuilding the history (\S\ref{sec:eval-hotpath}).
Capsules remain process-local and plan-specific because captured graphs contain
device addresses; they support compatible sessions within a process.
Integration defines the complete continuation state and valid resumption
points; the session manager copies the declared regions.

\section{Evaluation}
\label{sec:eval}
% \par{\small\color{red}\textbf{Kaiwen:} \textit{Suggestion from Codex; authors, please double-check.} Clarify the timing boundaries behind the main speedup claims. The appendix describes the 8.23\,ms result for $\pi_{0.5}$ on RTX 5090 as captured GPU replay; please verify that the 104.21\,ms eager baseline covers the same work before reporting the 12.66$\times$ speedup. State whether data transfers, preprocessing, and output processing are included, and distinguish GPU replay latency from full policy-call latency.\par}

%The evaluation follows the argument of \S\ref{sec:intro}. 
% \S\ref{sec:eval-e2e} compares \flashrt{} with today's deployment paths;
% \S\ref{sec:eval-fusion} and \S\ref{sec:eval-hotpath} examine pipeline composition,
% isolate launch overlap and value reuse, and report capsule costs; and
% \S\ref{sec:eval-fidelity} checks action fidelity and task success.
% Distribution to unmodified hosts (contribution 3): experiment design pending, see
% revision_plan_v6_eval.md, section 4.5. The former ladder figure (E5) is on hold with it.

\subsection{Setup}
\label{sec:eval-setup}
\label{sec:eval-method}
\textbf{GPUs.} We evaluate a Jetson AGX Thor (SM110, 128\,GB unified memory,
MAXN mode) for robot-side inference and an RTX 5090 (SM120, 32\,GB, fixed clocks)
for workstation inference. Both are Blackwell GPUs supporting FP8 and NVFP4.

\textbf{Policies.} Table~\ref{tab:models} lists six policies: two VLAs, one VA,
and three WAMs. All six run on RTX 5090; Thor covers $\pi_{0.5}$, GR00T N1.7,
LingBot-VA, and Cosmos3-Edge-Policy. Checkpoints and official code revisions are listed in the
supplementary material.

\begin{table}[t]
\centering
\caption{Policies evaluated. Steps: denoising steps per decision (video/action for LingBot-VA);
actions: actions returned per decision; views: cameras (Thor/RTX 5090 for GR00T N1.7).}
\label{tab:models}
\scriptsize
\setlength{\tabcolsep}{3pt}
\begin{tabular}{@{}llrrrr@{}}
\toprule
Class & Model & Params & Steps & Actions & Views \\
\midrule
VLA & $\pi_{0.5}$ & 3.35B & 10 & 10 & 2 \\
    & GR00T N1.7 & 3B & 4 & 16 & 1/2 \\
\midrule
VA  & LingBot-VA & 5.3B & 20/50 & 16 & 2 \\
\midrule
WAM & FLUX 3 Action & 7B & 4+CFG & 32 & 3 \\
    & Cosmos3-Edge-Policy & 4B & 4 & 32 & 3 \\
    & Motus & $\sim$8B & 10 & 16 & 3 \\
\bottomrule
\end{tabular}
\end{table}

\textbf{Baselines.} All baselines are official open-source implementations: OpenPI on Thor and
the existing LeRobot reference on RTX 5090 for $\pi_{0.5}$, Isaac-GR00T for GR00T N1.7, and the official releases of
LingBot-VA, FLUX 3 Action, Cosmos, and
Motus~\cite{openpi,lerobot,isaacgroot,lingbotva2026,flux3action2026,cosmos3_2026,motus2025},
each run eagerly and with its compiled or CUDA Graph path where one runs. Official
TensorRT deployments exist for $\pi_{0.5}$ and GR00T N1.7 on
Thor~\cite{jetsonopenpi,jetsongroot17}. Corresponding baseline paths were also
tested locally under aligned task protocols; published reference values are presented in Table~\ref{tab:e2e}.

\textbf{Protocol.} Experiments run offline on fixed or recorded inputs, excluding
physical action execution, network transfer, loading, calibration, and capture.
These measurements characterize warmed-up inference; control-loop timing
depends on the application's replanning schedule.
Comparisons use matching timing boundaries within each model--hardware pipeline;
host-specific intervals are described in Appendix~\ref{app:measurement}.
Paired mechanism experiments use
identical preprocessed inputs and pinned noise. Warm-up reaches up to 300 calls
on Thor to avoid a lower clock state. Table~\ref{tab:e2e} marks published
reference values; other latencies are measured locally. Compiled paths follow
the available official implementations.
Fidelity uses matched noise and the references specified in
Table~\ref{tab:fidelity}. No result uses approximate denoising-step caching such
as TeaCache~\cite{teacache2025}. Appendix~\ref{app:measurement} reports the
verified sampling and timing details for the controlled experiments.

\subsection{Deployment latency}
\label{sec:eval-e2e}
%\S\ref{sec:intro} argued that the deployment paths available today are slow.
Table~\ref{tab:e2e} compares \flashrt{} with the official eager, compiled, and
TensorRT paths, with published deployment references marked separately.
% Speedup is the Eager latency divided by the \flashrt{} latency.
% Precision mixes may differ, so this compares deployable systems, not fusion alone.
For $\pi_{0.5}$, \flashrt{} is 10.95$\times$ faster than official eager on Thor
and 12.66$\times$ faster on RTX 5090. GR00T N1.7 is 5.36$\times$ faster on
RTX 5090; its 24.39\,ms Thor result compares with the published eager reference
of approximately 126\,ms (about 5.2$\times$).
For LingBot-VA on Thor, a chunk takes 0.92\,s instead of 9.31\,s
(10.11$\times$), and on the RTX 5090 it is 5.72$\times$ faster. Among the WAMs, FLUX 3
Action and Motus are 4.91$\times$ and 7.10$\times$ faster on the RTX 5090, and
Cosmos3-Edge-Policy is 3.38$\times$ faster on Thor and 4.84$\times$ faster on the RTX 5090.
LingBot-VA/RTX 5090 uses FP8 video-step and INT4 weight-only action-step GEMMs,
with the same mixed-precision configuration in the latency and fidelity tables.
Appendix~\ref{app:measurement} specifies the operator precisions.
In the Compile column, the official \code{torch.compile} paths are 3.3\%
slower than eager for Cosmos/Thor and 2.2\% slower for LingBot-VA/RTX 5090.
LingBot-VA uses the official default compiler settings, which leave parts of
the pipeline in eager execution.

Published Thor TensorRT deployments report approximately 49\,ms for
$\pi_{0.5}$ and 39.9\,ms for GR00T N1.7, compared with \flashrt{}'s locally
measured 20.33 and 24.39\,ms, respectively.
% Published TensorRT provenance is stated in Setup and the table's dagger key.
For $\pi_{0.5}$/Thor, the deployment run has a 20.54\,ms p95 over 1000 timed
calls after 200 warm-ups, versus a 20.33\,ms median.
% Follow-up if existing logs are insufficient: sustained inference at the actual
% observation/action-chunk cadence, with latency percentiles and clock/thermal state.
% N1.6-3B official values, kept for reference: 117/105/92 ms (Thor), 58/37/31 ms (RTX 5090).
\begin{table*}[t]
\centering
\caption{Reported deployment latency (ms). Speedup is eager/\flashrt{} for
each model and GPU, combining pipeline and precision changes.
$\dagger$ marks published values and ratios computed from them;
other latencies are measured locally. --: no result.
Timing intervals: Appendix~\ref{app:measurement}.}
\label{tab:e2e}
\scriptsize
\setlength{\tabcolsep}{4pt}
\begin{tabular}{@{}llrrrrrrrrrr@{}}
\toprule
 & & \multicolumn{5}{c}{Jetson AGX Thor} & \multicolumn{5}{c}{RTX 5090} \\
\cmidrule(lr){3-7}\cmidrule(l){8-12}
Class & Model & Eager & Compile & TRT & \flashrt{} & Speedup & Eager & Compile & TRT & \flashrt{} & Speedup \\
\midrule
VLA & $\pi_{0.5}$ & 222.64 & 129.17 & $\sim49^{\dagger}$ & 20.33 & 10.95 & 104.21 & 58.26 & -- & 8.23 & 12.66 \\
    & GR00T N1.7 & $\sim126^{\dagger}$ & $\sim105^{\dagger}$ & $39.9^{\dagger}$ & 24.39 & $\sim5.2^{\dagger}$ & 41.10 & 32.01 & -- & 7.67 & 5.36 \\
\midrule
VA & LingBot-VA & 9309.53 & 6721.44 & -- & 920.9 & 10.11 & 2608.53 & 2666.62 & -- & 455.70 & 5.72 \\
\midrule
WAM & FLUX 3 Action & -- & -- & -- & -- & -- & 1353.57 & 1148.95 & -- & 275.51 & 4.91 \\
    & Cosmos3-Edge-Policy & 1223.00 & 1263.49 & -- & 362.15 & 3.38 & 586.28 & 514.93 & -- & 121.25 & 4.84 \\
    & Motus & -- & -- & -- & -- & -- & 1186.10 & 1403.15 & -- & 167.08 & 7.10 \\
\bottomrule
\end{tabular}
\par\vspace{2pt}\scriptsize
Host timing intervals, mixed-precision settings, and sampling are detailed
in Appendix~\ref{app:measurement}.
\end{table*}

\subsection{Deep fusion and launch overlap}
\label{sec:eval-fusion}
Observations 1 and 2 (\S\ref{sec:motivation}) identify intermediate-kernel
costs and independent work across dependent kernels. Deep kernel fusion removes
boundaries; launch overlap may hide part of the cost at the boundaries that
remain. Table~\ref{tab:ablation} reports latency at each precision tier,
Table~\ref{tab:kernels} counts kernels per decision, and
Figure~\ref{fig:deep-fusion} gives a measured layer trace and an isolated PDL
off/on comparison.

\textbf{Precision tiers.} Table~\ref{tab:ablation} compares the official compiled
path with \flashrt{} at 16, 8, and 4 bits. Already at 16 bits, without lower
precision, \flashrt{} is 1.3--3.7$\times$ faster than the official compiled path.
This deployment comparison combines specialized kernels, fusion, graph replay,
and value reuse (\S\ref{sec:eval-hotpath}). The table compares pipeline
configurations at each precision; the 4-bit entries match Table~\ref{tab:e2e}.
Precision, layout, fusion, and launch scheduling are co-designed: changing one
can require conversions or a different kernel schedule. We therefore interpret
these rows as complete configurations, rather than independent gains to add.
The paired PDL and reuse experiments hold the accompanying implementation
fixed to measure those changes within a valid pipeline.
% Figure 3 provides measured structural evidence for the co-designed layer;
% an additional fusion-only experiment is not required for this framing.

\begin{table*}[t]
\centering
\caption{Latency (ms) across pipeline configurations at 16, 8, and 4 bits.
The 16-bit comparison matches FP16/BF16 precision; lower tiers change precision
and kernels. Baseline sources follow Table~\ref{tab:e2e}.}
\label{tab:ablation}
\scriptsize
\setlength{\tabcolsep}{4pt}
\begin{tabular}{@{}lrrrrrrr@{}}
\toprule
 & \multicolumn{2}{c}{$\pi_{0.5}$} & \multicolumn{2}{c}{GR00T N1.7} & \multicolumn{2}{c}{LingBot-VA} & FLUX 3 Action \\
\cmidrule(lr){2-3}\cmidrule(lr){4-5}\cmidrule(lr){6-7}\cmidrule(l){8-8}
Configuration & Thor & 5090 & Thor & 5090 & Thor & 5090 & 5090 \\
\midrule
Official compile, 16-bit & 129.17 & 58.26 & $\sim$105 & 32.01 & 6721.44 & 2666.62 & 1148.95 \\
\flashrt{} 16-bit & 67.47 & 25.08 & 54.77 & 18.85 & 3174.26 & 726.10 & 900.34 \\
\flashrt{} 8-bit & 40.71 & 13.85 & 36.98 & 12.91 & 2029.42 & 532.69 & 423.92 \\
\flashrt{} 4-bit & 20.33 & 8.23 & 24.39 & 7.67 & 920.9 & 455.70 & 275.51 \\
\midrule
16-bit vs.\ 16-bit compile & 1.91$\times$ & 2.32$\times$ & $\sim$1.9$\times$ & 1.70$\times$ & 2.12$\times$ & 3.67$\times$ & 1.28$\times$ \\
4-bit vs.\ 16-bit compile & 6.35$\times$ & 7.08$\times$ & $\sim$4.3$\times$ & 4.17$\times$ & 7.30$\times$ & 5.85$\times$ & 4.17$\times$ \\
\bottomrule
\end{tabular}
\end{table*}

\textbf{Fusion.} Table~\ref{tab:kernels} counts the GPU kernels in one decision.
The complete \flashrt{} pipelines run 59--93\% fewer kernels than official
eager paths; the reduction includes fusion and removal of repeated work. In
the layer trace of Figure~\ref{fig:deep-fusion}, the \code{torch.compile}+FP8 path
takes 193.2\,$\mu$s per layer (20 kernels), while the \flashrt{} FP8 path takes
105.2\,$\mu$s (12 kernels). Both use FP8 GEMMs; the \flashrt{} layer combines
fewer intermediate boundaries with specialized GEMM implementations. This
comparison measures the co-designed layer. The NVFP4 path takes 62.1\,$\mu$s
with 11 kernels, combining its precision choice with a fused GeGLU epilogue.

\begin{table}[t]
\centering
\caption{GPU kernels per decision of the official eager implementation and \flashrt{},
counting every kernel inside CUDA Graphs.}
\label{tab:kernels}
\scriptsize
\setlength{\tabcolsep}{4pt}
\begin{tabular}{@{}llrrr@{}}
\toprule
Policy & GPU & Eager & \flashrt{} & Reduction \\
\midrule
$\pi_{0.5}$ & Thor & 13{,}414 & 2{,}610 & 81\% \\
 & RTX 5090 & 13{,}134 & 2{,}216 & 83\% \\
GR00T N1.7 & Thor & 5{,}190 & 2{,}148 & 59\% \\
 & RTX 5090 & 4{,}978 & 1{,}579 & 68\% \\
LingBot-VA & Thor & 180{,}245 & 27{,}089 & 85\% \\
 & RTX 5090 & 220{,}504 & 35{,}557 & 84\% \\
FLUX 3 Action & RTX 5090 & 18{,}759 & 1{,}268 & 93\% \\
\bottomrule
\end{tabular}
\end{table}

\textbf{Launch overlap.} The bottom two rows of Figure~\ref{fig:deep-fusion}
isolate PDL on Thor: the same NVFP4 path replays the same 11 kernels with PDL off
and then on at the FFN and adjacent O/QKV boundaries. The layer period falls from
62.1 to 52.1\,$\mu$s, and the FFN window within it falls from 33.92 to
27.30\,$\mu$s. In an unprofiled measurement of the whole decoder, enabling PDL at
these boundaries lowers the median from 10.124 to 9.207\,ms, and enabling it at all
decoder boundaries lowers it to 9.009\,ms. With PDL at all decoder boundaries, the
full call falls from 21.960 to 21.044\,ms. All paths run the same kernels at the
same precision, a 4.2\% whole-call gain in the paired PDL experiment
(Appendix~\ref{app:measurement}). On the RTX 5090,
PDL shortens one FFN block of the action expert
from 14.56 to 12.09\,$\mu$s. The whole-call gain is smaller because launch overlap
changes only eligible action-expert boundaries; it does not shorten the other
stages or host work.

\begin{takeaway}{Fast kernels can still form a slow pipeline.}
Separate kernels still move intermediate data and wait for dependencies.
Fusion removes selected boundaries; launch overlap reduces waiting at
those that remain.
\end{takeaway}

\subsection{Frozen execution state}
\label{sec:eval-hotpath}
Observation 3 (\S\ref{sec:motivation}) showed that values needed by later decisions
are often recomputed. Frozen execution state makes these values explicit: the execution
contract names their buffers, owners, and refresh rules. This lets \flashrt{} retain
unchanged values and copy declared state regions into an execution-state capsule at a
stage boundary. Table~\ref{tab:state} summarizes these two uses.

\begin{table}[t]
\centering
\caption{Benefits of frozen execution state (ms): reusing values across decisions and
resuming a session after another session overwrites its buffers, with the cost of capsule
copies. 
%Restored and reused actions match the reference bit for bit.
Paired experiment protocol: Appendix~\ref{app:measurement}.}
\label{tab:state}
\scriptsize
\setlength{\tabcolsep}{2pt}
\begin{tabular}{@{}llrrrr@{}}
\toprule
Use & Policy/GPU & Baseline & Frozen state & Speedup \\
\midrule
Reuse values & $\pi_{0.5}$/Thor & 37.05 & 21.61 & 1.71$\times$ \\
Resume after overwrite & $\pi_{0.5}$/Thor & 21.48 & 9.73 & 2.21$\times$ \\
Resume after overwrite & LingBot-VA/Thor & 1809.32 & 925.27 & 1.96$\times$ \\
\bottomrule
\toprule
\multicolumn{2}{@{}l}{\emph{Capsule copy cost}} & MB & Snapshot & Restore & Fork \\
\midrule
$\pi_{0.5}$ & Thor & 14.40 & 0.48 & 0.13 & 0.50 \\
LingBot-VA & Thor & 1410.65 & 18.77 & 11.29 & 36.03 \\
\bottomrule
\end{tabular}
\end{table}

\textbf{Reuse across decisions.} We run $\pi_{0.5}$ on Thor with the deployed NVFP4
engine over eight recorded observations of one task. Both paths replay the same
captured graphs; the baseline also recomputes the prompt embedding and the time
conditioning of every denoising step into the same buffers. Keeping these values
reduces the mean decision latency from 37.05 to 21.61\,ms, and the actions stay
bit for bit the same. The baseline uses the existing preparation code rather than a
separately optimized implementation.

\textbf{Restore after an interruption.} Persistent values can remain in place while one
session runs uninterrupted. When sessions share captured graphs and fixed working
buffers, however, another session may overwrite them. At a stage boundary,
\code{snapshot} copies the declared state regions into an execution-state capsule;
\code{restore} copies them back before replay resumes. If each session permanently
retains its own working buffers, no restore is needed. Capsules serve suspended
sessions or rollout branches that share one working set: a saved branch retains
declared state without duplicating the entire workspace. Table~\ref{tab:state}
reports the state footprint and snapshot, restore, and fork times.
% Follow-up: end-to-end multi-session and branching-workload evaluation.

For $\pi_{0.5}$, we interrupt a session after its context stage, run another session over
the shared buffers, and then resume the original session. Rebuilding its instruction-
and image-conditioned state and generating the action takes 21.48\,ms. Restoring its
14.40\,MB capsule and generating the same action takes 9.73\,ms. The saved context
computation was completed before the interruption; a fresh image would still require
a new context stage.

\begin{figure}[t]
\centering
\begin{tikzpicture}
\begin{semilogyaxis}[
  width=0.8\columnwidth, height=4.4cm,
  xmode=log, log basis x=2,
  xtick={1,2,4,8,16}, xticklabels={1,2,4,8,16},
  xlabel={History length (chunks)}, ylabel={Time until state is ready (ms)},
  ymin=5, ymax=60000,
  tick label style={font=\scriptsize}, label style={font=\scriptsize},
  legend columns=3, legend cell align=left,
  legend style={font=\scriptsize, draw=none, fill=none, at={(0.5,1.03)}, anchor=south},
  grid=major, grid style={gray!20}]
\addplot[mark=square*, color=red!70!black] coordinates
  {(1,870.0) (2,1696.7) (4,3381.1) (8,6872.1) (16,14405.3)};
\addlegendentry{Rerun past chunks}
\addplot[mark=triangle*, color=orange!80!black] coordinates
  {(1,79.9) (2,132.3) (4,237.8) (8,451.9) (16,894.7)};
\addlegendentry{Rebuild K/V}
\addplot[mark=*, color=blue!70!black] coordinates
  {(1,11.23) (2,11.28) (4,11.17) (8,11.21) (16,11.29)};
\addlegendentry{Restore}
\end{semilogyaxis}
\end{tikzpicture}
\vspace{-10pt}
\caption{LingBot-VA on Thor: time until the history state is ready again after another
session overwrites it (log scale, median of 10 runs). Generating the next chunk
(0.78--0.92\,s) is not included. 
%Restore uses the same state-ready interval as Table~\ref{tab:state}: 11.29\,ms at 16 chunks.
}
\label{fig:capsule}
\end{figure}
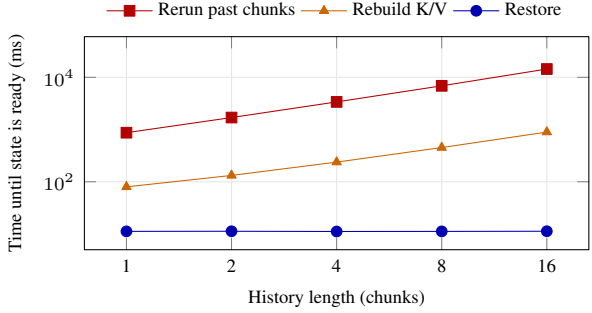

LingBot-VA carries history across action chunks. After another session overwrites its
working buffers, our rebuild baseline re-encodes the saved observations and actions to
reconstruct the historical K/V, while skipping the earlier denoising computations. At
16 committed chunks, this reconstruction takes 894.71\,ms, whereas restoring the
capsule takes 11.29\,ms. Both paths must then generate the next action chunk. Including
that work, time to the next action is 1809.32\,ms after reconstruction and 925.27\,ms
after restoration, a 1.96$\times$ speedup. Forking the LingBot-VA state into a new
session takes 36.03\,ms. Figure~\ref{fig:capsule} shows that K/V
reconstruction grows from 79.88\,ms after one chunk to 894.71\,ms after 16, while
restoration remains about 11.2\,ms because the current implementation copies a fixed
allocated state region. Rerunning all past chunks is more expensive still, reaching
14.4\,s at 16 chunks.

\begin{takeaway}{Useful state can outlive an inference call.}
Prepared values remain reusable until their inputs change. After an
interruption, restoring saved state can avoid rebuilding completed context.
\end{takeaway}

\subsection{Fidelity and task success}
\label{sec:eval-fidelity}
Table~\ref{tab:fidelity} reports matched-input, matched-noise action agreement
against the official implementation, except for the FP16-frontend references
marked $^{d}$. LingBot-VA/RTX 5090 is evaluated on held-out traces without
sample filtering. Its six arm channels are reported here; full-action
scores including the gripper appear in Appendix~\ref{app:lingbot-fidelity}.
On LIBERO-10, $\pi_{0.5}$ on \flashrt{} with 4-bit precision succeeds in
94.04\% of trials on Thor and 93.32\% on RTX 5090,
%\footnote{The RTX 5090 result aggregates 4,666 successes over 5,000 trials using ten seeds.} 
compared with 92.4\% for
the measured official FP16 reference. LingBot-VA's evaluated FP8/INT4 path
succeeds in 97/100 episodes on RTX 5090, versus 98/100 for official BF16,
using matched seeds and initial states across ten LIBERO-10 tasks.
% Follow-up: collect existing reference-run trial/seed metadata and existing
% task-validation summaries across models; statistical superiority is not a claim.
% These simulation success rates are behavioral checks, not safety certification.

\begin{table}[t]
\centering
\caption{Action agreement: global / worst-sample cosine under matched noise.
$n$ marks per-input scores; Motus gives the range over four inputs.
$^{c}$: six arm channels, 717 chunks, official
FP32 reference; full actions: Appendix~\ref{app:lingbot-fidelity}.
$^{d}$: FP16 frontend reference. Bottom: LIBERO-10 success (100 matched
episodes for LingBot-VA). Protocols: Appendix~\ref{app:measurement}.}
\label{tab:fidelity}
\scriptsize
\setlength{\tabcolsep}{3pt}
\begin{tabular}{@{}llrr@{}}
\toprule
Class & Model & Thor & RTX 5090 \\
\midrule
VLA & $\pi_{0.5}$ & .999713 / .999360 & .999850 / .999201$^{d}$ \\
    & GR00T N1.7 & .9997 ($n{=}1$) & .99999$^{d}$ ($n{=}1$) \\
VA  & LingBot-VA, BF16 & .999949 / .999920 & .999502 / .952046$^{c}$ \\
    & LingBot-VA, 4-bit mix & .995361 / .993777 & .998907 / .941478$^{c}$ \\
WAM & FLUX 3 Action & -- & .999850 / .998939 \\
    & Cosmos3-Edge-Policy & .998533 / .976768 & .998138 / .988953 \\
    & Motus ($n{=}4$) & -- & .99991--.99994 \\
\midrule
\multicolumn{2}{@{}l}{LIBERO-10 success (\%)} & Thor & RTX 5090 \\
\multicolumn{2}{@{}l}{$\pi_{0.5}$, official FP16} & \multicolumn{2}{c}{92.4} \\
\multicolumn{2}{@{}l}{$\pi_{0.5}$, \flashrt{} 4-bit} & 94.04 & 93.32 \\
\multicolumn{2}{@{}l}{LingBot-VA, official BF16} & -- & 98.0 \\
\multicolumn{2}{@{}l}{LingBot-VA, FP8/INT4} & -- & 97.0 \\
\bottomrule
\end{tabular}
\end{table}

% ===========================================================================
% \section{Deployment and Industrial Impact}
% \label{sec:impact}

\section{Deployment}
\label{sec:structures}
\label{sec:modes}
\flashrt{} offers native C++, a Python frontend, and selective PyTorch-host
replacement. All use its kernels; captured-stage exports share the execution
contract (\S\ref{sec:exec}). Pipeline ownership determines the optimization scope.

\textbf{Native engine and frontend.} The native engine owns the pipeline,
fixed buffers, and captured graphs. The host fills swap windows and reports
changed inputs; the session manager schedules replay and capsule copies.
C++ avoids per-kernel host dispatch. The Python frontend uses the same
kernels and stages for integration and evaluation, with Python orchestration,
input preparation, and output conversion.

\textbf{Existing PyTorch hosts.}\label{sec:distribute}
Each replaceable region has a specification and PyTorch reference. Qualification
checks inputs, outputs, state, shapes, and loop count; tensor-dependent host
branches prevent capture. After host-side calibration, activation requires
reference agreement and lower whole-call latency, otherwise retaining the
original path. Kernels ship through the Kernel Hub. Graph-safe stages can use
\code{capture(fn, windows)} and export replay and capsule interfaces. The host
retains its control loop; local edits may remove synchronization or enable
capture. Host module boundaries limit fusion, and preprocessing, postprocessing,
and tensor creation remain outside capture.

\textbf{Integration work.} Engineers specify stage boundaries, buffer lifetimes,
precision/layout interfaces, and kernel variants, then validate the complete
policy. Loading, calibration, and graph capture execute the specified setup;
they do not automatically discover this design. Shared kernels and contract
interfaces reduce repeated implementation, but a new model--hardware pipeline still
requires model-specific design and validation.
% Follow-up: quantify person-weeks and model-specific code size.

\begin{takeaway}{Integration can erase kernel speedups.}
Extra conversion and synchronization can offset a faster kernel. Evaluate
replacements within the complete application path, preserving output
agreement and the existing control loop.
\end{takeaway}
% Follow-up: qualification counts, gate overhead, and fallback rates.

\section{Ecosystem Adoption and Impact}
\label{sec:impact}
\flashrt{} is an open-source inference engine (\url{https://github.com/flashrt-project/FlashRT}),
with kernels on the Hugging Face Kernel Hub. The October 2026 repository
snapshot records 658 GitHub stars and 89 forks, alongside the public
integrations in Table~\ref{tab:adoption}.
Its public uses span robot-model developers, host frameworks, and vendor
deployment guides. Beyond NVIDIA GPUs, it supports AMD Instinct and Ascend;
an external team extended the AMD backend from MI350X
to MI300X. Table~\ref{tab:adoption} lists uses documented publicly as of October
2026; latency figures are reported by the adopters under their respective setups.

\begin{table}[t]
\centering
\caption{Public uses of \flashrt{} as of October 2026. Links beside project
names open the corresponding public sources.}
\label{tab:adoption}
\scriptsize
\setlength{\tabcolsep}{4pt}
\newcommand{\adoptionlink}[1]{\,\href{#1}{[link]}}
\begin{tabular}{@{}ll@{}}
\toprule
Project & How it uses \flashrt{} \\
\midrule
MindOn Mind-1\adoptionlink{https://www.mindon.tech/blog/mind-1/index.html} & Inference backend for a robot model \\
EagleVLA-Edge\adoptionlink{https://github.com/PKU-SEC-Lab/EagleVLA-Edge} & Edge VLA engine that plugs into \flashrt{} \\
LeRobot\adoptionlink{https://github.com/huggingface/lerobot/pull/3870} & Optional FP8 kernels for $\pi_{0.5}$ \\
FineART-VLA\adoptionlink{https://arxiv.org/abs/2609.36416} & Optional kernels for training and inference \\
Jetson AI Lab\adoptionlink{https://www.jetson-ai-lab.com/tutorials/openpi_on_thor/} & TensorRT guide for $\pi_{0.5}$, inspired by \flashrt{} \\
APXInf\adoptionlink{https://github.com/RLinf/APXinf-robo} & Edge inference engine that credits \flashrt{} \\
\bottomrule
\end{tabular}
\end{table}

MindOn reports 82 to 32\,ms mean latency for Mind-1 using \flashrt{}~\cite{mindon2026}.
LeRobot's separate PR setup reports 89.4 to 46.7\,ms for opt-in $\pi_{0.5}$
on RTX 5090: a bit-exact denoising loop reaches 79.3\,ms, followed by FP8
feed-forward kernels from the Kernel Hub~\cite{lerobotpr3870,hfkernels}.
The option is disabled by default with BF16 fallback. Jetson AI Lab credits
\flashrt{}'s Thor results as motivation for its $\pi_{0.5}$ TensorRT
deployment~\cite{jetsonopenpi}, used as our published baseline.
% Independently remeasure LeRobot option-on/off under the paper's end-to-end protocol.

% ===========================================================================
\section{Related Work}
\label{sec:related}
\textbf{Graph compilers.} TVM, TensorRT, and \code{torch.compile} provide
fusion, layouts, autotuning, and CUDA Graph replay~\cite{tvm2018,tensorrt,pytorch2,cudagraphs}.
\flashrt{} supplies model--hardware precision/layout interfaces and explicit
cross-decision state rules, then records the resulting pipeline in a contract;
compiler-generated regions can use the same interface.

\textbf{Reuse, serving, and model-side compute.} PagedAttention, SGLang, Prompt
Cache, and Hydragen share or reuse text-prefix attention
state~\cite{paged2023,sglang2024,promptcache2024,hydragen2024}; Marconi manages
hybrid KV/recurrent prefix reuse and HCache accelerates contextual KV
restoration~\cite{marconi2025,hcache2025}. Separately, token
pruning~\cite{efficientvla2025}, approximate feature
caching~\cite{vlacache2025,teacache2025}, and fewer-step
distillation~\cite{consistencypolicy2024,flashwam2026} reduce policy computation,
while real-time chunking~\cite{rtc2025} overlaps inference with action execution.
\flashrt{} applies reuse to instruction, step-conditioning, K/V, VAE, and RNG
regions with different update rates in a robot loop, under explicit ownership and
valid resumption boundaries; it avoids recomputing values whose inputs are
unchanged and is complementary to these model-side and scheduling methods.

\textbf{Kernels and LLM runtimes.} CUTLASS and FlashAttention supply GEMM and
attention kernels~\cite{cutlass,flashattention2}; FlashInfer adds customizable,
graph-compatible attention~\cite{flashinfer2025}, Marlin optimizes FP16$\times$INT4
GEMM~\cite{marlin2024}, and TensorRT-LLM integrates optimized kernels and LLM
request scheduling~\cite{tensorrtllm}; these libraries can implement \flashrt{}'s
stages.

\textbf{VLA inference systems and vendor deployments.} \citet{realtimevla2025}
reduce inference overheads for 30\,Hz $\pi_0$-class execution; BLURR combines
prefix caching, mixed precision, and single-step rollout~\cite{blurr2025};
EagleVLA combines a llama.cpp engine with foresight-aligned asynchronous
inference on Orin~\cite{eaglevlacorl2026}, and its engine is a \flashrt{}
provider~\cite{eaglevla2026}; Reflex adds batched visual encoding and graph
replay~\cite{reflex2026}; FlashDrive co-designs a driving VLA's algorithm and
system~\cite{flashdrive2026}. NVIDIA supplies $\pi_{0.5}$ and GR00T TensorRT
paths~\cite{jetsonopenpi,grootdeploy} and reports Cosmos3-Edge-Policy at 15\,Hz
on Thor~\cite{cosmospolicyblog}; DreamZero combines two-GPU parallelism,
compilation, kernels, quantization, and caching~\cite{dreamzero2026}. Some of
these change model computation, and all require model-specific integration;
\flashrt{} preserves architecture and denoising schedule across VLA, VA, and
WAM pipelines and shares contract and kernel interfaces, with public
EagleVLA-Edge and APXInf integrations listed in Table~\ref{tab:adoption}.

% ===========================================================================
\section{Limitations and Conclusion}
\label{sec:conclusion}
\textbf{Limitations.} \flashrt{} targets small-batch, fixed-shape inference; new
shapes need recapture, and capsules resume compatible sessions within one process
(\S\ref{sec:capsule}). The current evaluation covers warmed-up inference on two
Blackwell GPUs and task success in LIBERO simulation.
% Follow-up: summarize existing validation; paced long-run characterization if
% needed. Host qualification and multi-session workloads are extensions, not
% required experiments for the current single-robot inference scope.

\textbf{Conclusion.} Fast kernels are necessary but not sufficient.
\flashrt{} co-designs their interfaces, launch dependencies, and state
lifetimes, recording the result in an execution contract. Across four policies
on Thor and six on RTX 5090, evaluated paths are 3.38--12.66$\times$ faster
than official eager paths and 1.3--3.7$\times$ faster than compiled paths at
matched 16-bit precision. %MindOn and LeRobot illustrate native execution and selective adoption
\flashrt{} has delivered tangible impact across industry and the open-source
research community (\S\ref{sec:impact}): it is open source, its kernels are
distributed through the Hugging Face Kernel Hub, and its opt-in kernels have
been adopted in LeRobot.

Next, we aim to automate pipeline optimization so new physical-AI models
can run efficiently across diverse hardware with minimal manual effort.

% Incoming conclusion wording retained for reference:
% FlashRT has already delivered tangible impact (\S\ref{sec:impact}) across both industry and the open-source research community: it is fully open source, distributed via the Hugging Face Kernel Hub, and its opt-in kernels have been merged into mainline LeRobot.
%
% Looking ahead, we envision FlashRT evolving into the standard runtime foundation for real-time physical AI. More broadly, we believe the pipeline co-design paradigm will replace ad-hoc, per-model performance tuning as the standard approach to deploying production physical AI systems.

% ==========================================================================
\bibliographystyle{plainnat}
\bibliography{refs_flashrt}

\section{AI Usage Disclosure}
% No AI tools were used in writing this paper. AI coding agents were used in parts of the development of the open-source FlashRT project. The extent of their use can be verified through the project's commit history.
AI Usage Disclosure. AI assistants (ChatGPT, specifically) were only used to help refine the manuscript's grammar and assist authors with writing suggestions. Separately, in developing the open-source FlashRT code (verifiable in the project's public commit history). All content was reviewed and is the responsibility of the authors. Literature search, experiment design, data collection, and results were produced by the authors, no content or citations were AI-generated.

\clearpage
\appendix
\section{Worked Examples}
\label{app:examples}

The worked examples below show the contract and verified measurement details,
then the deployment facts behind fusion, launch overlap, and state reuse.

\subsection{Worked Execution Contract}
\label{app:contract}
The contract is exported through a versioned C interface. It lists captured
graphs and their supported shapes, named buffers, restorable byte regions,
and an identity fingerprint used to match saved state to its execution plan.
The model interface adds inputs with shapes and update methods, plus stages
with explicit prerequisites.

\textbf{Execution rules.} A retained value can be reused only while its source
inputs are unchanged. Stages consume prerequisite outputs valid for the
current inputs. Restored state belongs to a compatible plan and resumes at
the declared boundary; a new observation still requires a new context stage.
These rules specify correct use of the interface. Listing~\ref{lst:contract-fields}
shows two of its records; names and field types follow the implementation.

\begin{lstlisting}[language=C, captionpos=b,
caption={Contract records for saved regions and stage order. A region names
the buffer bytes to copy; a stage names its graph and prerequisite stages.},
label={lst:contract-fields}]
typedef struct frt_runtime_region_desc {
  const char* name;
  frt_buffer buffer;
  uint64_t offset, bytes;
  uint32_t flags, reserved;
} frt_runtime_region_desc;

typedef struct frt_runtime_stage_desc {
  uint32_t graph;
  uint32_t n_after;
  const uint32_t* after;
} frt_runtime_stage_desc;
\end{lstlisting}

\textbf{Engine checks and input updates.} Construction rejects stage references
to missing graphs or prerequisites that have not been declared. An input that
requires processing must provide its update function. Replay requires a captured
variant for the requested shape, and restore checks state-layout and plan
compatibility. For example, the GR00T N1.7 Thor frontend rejects a second prompt
initialization and requires a new frontend instance for a new prompt.

\textbf{Caller responsibilities.} The normal model call performs its declared
input processing and stage sequence. A host driving individual stages must
submit current inputs and refresh dependent state before action replay. If it
omits an update or reuses context from an older observation, the result can be
an action based on outdated inputs. Shape and plan checks do not detect such
unreported changes. The interface checks execution compatibility; the host
retains responsibility for input freshness and the robot's control loop.

\textbf{Worked plan.} In the split $\pi_{0.5}$ plan below, the engine owns the
fixed allocations and replay handles; the application supplies input updates,
and the session manager orders stages and capsule copies.

\begin{table}[h]
\centering
\caption{Buffer roles and update events in the split $\pi_{0.5}$ plan.}
\scriptsize
\setlength{\tabcolsep}{3pt}
\begin{tabular}{@{}lll@{}}
\toprule
Buffer & Role / writer & Refresh event \\
\midrule
Images & Swap / host & Fresh observation \\
Noise & Swap / host & New action generation \\
Prompt embedding & Static / preparation & Prompt change \\
Step conditioning & Static / preparation & Schedule change \\
Context K/V & State / context & Image or prompt change \\
Actions & Swap / action & Action replay \\
\bottomrule
\end{tabular}
\end{table}

Preparation writes instruction embeddings and fixed-step conditioning. Context
reads images and the prepared prompt and writes K/V. Action depends on context,
reads that K/V and the noise window, and produces the action chunk. A fresh
observation therefore invalidates context, even when the instruction is
unchanged. Each replay plan is selected by its supported shape; changing a
shape requires preparation and capture before replay.

At the post-context boundary, the measured capsule contains context K/V,
instruction embeddings, and the decoder position table, totaling 14.40\,MB.
The exported region list gives names, offsets, and lengths; schema and plan
fingerprint checks restrict restore to a compatible session. Weights and fixed
step tables remain resident and are not copied in this experiment. Restoring
this state permits continuation from the completed context; a new image
refreshes the context stage. The application supplies the input-change events
that determine when retained values are refreshed.

\subsection{Measurement Details and Short-Run Distributions}
\label{app:measurement}
Timing boundaries are specific to each host. The deployed $\pi_{0.5}$/Thor
\code{infer} interval includes image upload, model execution, action transfer,
and decoding, excluding prompt setup. The 8.23\,ms
$\pi_{0.5}$/RTX 5090 entry measures captured GPU replay. Cosmos/Thor measures
in-process CPU observation to CPU action. GR00T/RTX 5090 measures GPU \code{infer(aux)},
including fresh vision features and the complete DiT.
LingBot-VA/Thor reports the mean over saturated chunks 43--47; its RTX 5090
measurement covers history update and prediction, excluding reset, network
transfer, and text encoding. Cosmos/RTX 5090 includes preprocessing, VAE and
text processing, denoising, and action formatting in both paths. Published
reference ratios retain their original source protocols; they are distinct
from the paired PDL and reuse experiments.

The controlled $\pi_{0.5}$/Thor state experiment uses engine source
\code{e4bd927}, PyTorch 2.14.0+cu130, and CUDA 13.0 in a separate host
environment from the deployment image. Reuse averages 12 alternating pairs
over eight recorded observations; both sides replay identical graphs and
differ only in recomputing prompt embeddings and step conditioning. Resume,
snapshot, and restore-only statistics use 64 paired trials. Every restored
continuation and reuse comparison is bit-exact. The PDL experiment uses the
same engine source and kernel inventory, 30 warm-ups and 200 interleaved
CUDA-event decoder samples per mode; synchronized whole-call medians span
eight recorded observations, with 40 bit-exact gates.

\begin{table}[h]
\centering
\caption{Short-run Thor latency distributions (ms). Rows retain their own
timing intervals; decoder p95 is not whole-call p95.}
\scriptsize
\setlength{\tabcolsep}{3pt}
\begin{tabular}{@{}lrrr@{}}
\toprule
Interval & Samples & p50 & p95 \\
\midrule
Deployed $\pi_{0.5}$ infer & 1000 & 20.33 & 20.54 \\
PDL off, decoder & 200 & 10.124 & 10.137 \\
PDL FFN + O/QKV, decoder & 200 & 9.207 & 9.220 \\
PDL all, decoder & 200 & 9.009 & 9.024 \\
Rebuild + action & 64 & 21.484 & 21.533 \\
Restore + action & 64 & 9.726 & 9.778 \\
Snapshot only & 64 & 0.477 & 0.531 \\
Restore only & 64 & 0.131 & 0.139 \\
\bottomrule
\end{tabular}
\end{table}

The deployment run uses 200 warm-ups; the controlled state and PDL results
report paired changes under the stated configuration. Their 21.61\,ms reuse
and 21.960\,ms PDL-off intervals were collected in the controlled host
environment; the 20.33\,ms deployment interval comes from the deployment image.
The distributions summarize the stated warmed-up runs. Sampling counts and
compiler settings are recorded per configuration: Cosmos/Thor uses default
\code{torch.compile} settings and 60 timed calls per
deployment path, with p90 of 1227.45, 1267.09, and 363.86\,ms for eager,
compile, and mixed 4-bit respectively.
Table~\ref{tab:ablation} compares configurations collected across precision
tiers and measurement sessions; controlled mechanism changes are measured by
the paired PDL and reuse experiments. Capsule results measure state-copy and
continuation operations over shared working buffers.
% Deployment latency characterization should use the actual periodic request
% cadence and include time-to-action, clocks, temperature, and throttling.
% An uninterrupted maximum-rate stress test changes duty cycle and is a separate
% workload. Reuse existing paced logs before commissioning another run.

\textbf{Fidelity samples.} $\pi_{0.5}$ uses eight observations on Thor and 24
real LIBERO frames on RTX 5090; the latter has 16 separate calibration frames.
GR00T/Thor uses one single-view observation. The RTX 5090 GR00T result is one
two-view LIBERO observation at four denoising steps, compared with the FP16
frontend; its 0.999985 cosine matches the workload of the 7.67\,ms latency
entry. Cosmos uses 114 requests (38
observations, three noise seeds) on Thor and 60 requests (12 observations,
five seeds) on RTX 5090. Motus uses four RoboTwin2 input bundles
(Appendix~\ref{app:motus-fidelity}). LingBot-VA/RTX 5090 scores
717 chunks from 40 traces held out from calibration, against the official
FP32 reference with TF32 disabled. Arm and full-action results are reported below. The
evaluated mixed-precision configuration uses FP8 for the video backbone's cross-attention query/output,
self-attention QKV, and FFN output projections, and INT4 weight-only GEMMs for
action steps; attention operations and history updates retain their
higher-precision paths. It retains 20 video and 50 action denoising steps.
This configuration is timed at 455.70\,ms and is used for both the filtered
diagnostic and the unfiltered results below.
For $\pi_{0.5}$, LIBERO success rates compare the measured official FP16
reference with the 4-bit \flashrt{} rollouts. LingBot-VA uses its official
BF16 reference and the FP8/INT4 configuration described above. Its paired
evaluation runs tasks 0--9 and episodes 0--9 with the same seeds and initial
states: 98/100 successes for the reference and 97/100 for \flashrt{}.
% Follow-up: archive reference-run metadata and existing task-validation
% summaries. Reference provenance and existing baseline validation clarified by SL.

\textbf{Validation represented in the tables.}
Table~\ref{tab:fidelity} reports returned-action comparisons for all six
policies, and closed-loop success rates for $\pi_{0.5}$ and LingBot-VA. The accompanying
records cover the following inputs and evaluation settings.

\begin{table}[t]
\centering
\caption{Recorded validation by policy.}
\scriptsize
\setlength{\tabcolsep}{3pt}
\begin{tabular}{@{}lp{0.59\columnwidth}@{}}
\toprule
Policy & Recorded validation \\
\midrule
$\pi_{0.5}$ & Recorded-input action comparisons; LIBERO-10 rollouts against the measured FP16 reference. \\
GR00T N1.7 & Single-view Thor and two-view RTX 5090 LIBERO observations; single-observation cosine. \\
LingBot-VA & RTX 5090: held-out LIBERO traces for arm/full-action agreement; paired 100-episode LIBERO-10 success comparison. \\
FLUX 3 Action & Returned-action agreement against the official implementation under matched inputs and noise. \\
Cosmos3-Edge-Policy & DROID observations: 114 observation/noise requests on Thor and 60 on RTX 5090. \\
Motus & Four RoboTwin2 input bundles; action and decoded-video agreement against saved upstream outputs. \\
\bottomrule
\end{tabular}
\end{table}
% SL reports additional completed task-level validation for all models. Add its
% environments, tested configurations, metrics, and results once corresponding
% summaries are available; do not infer rollout success from replay cosine.

\subsubsection{LingBot-VA: unfiltered action agreement}
\label{app:lingbot-fidelity}
Table~\ref{tab:lingbot-unfiltered} reports all recorded chunks, including
numerically sensitive ones. Each chunk contains 16 actions with six arm
channels and a binary gripper command. We report arm motion separately from
the complete action because a change in gripper-switch timing can strongly
affect the full-action cosine.

\begin{table}[h]
\centering
\caption{Unfiltered LingBot-VA/RTX 5090 action agreement against official
FP32. Entries are global / worst-chunk cosine. The mixed configuration
matches the 455.70\,ms deployment.}
\label{tab:lingbot-unfiltered}
\scriptsize
\setlength{\tabcolsep}{3pt}
\begin{tabular}{@{}lrr@{}}
\toprule
Configuration & Six arm channels & Full action \\
\midrule
BF16 & .999502 / .952046 & .997388 / .643595 \\
FP8/INT4 mix & .998907 / .941478 & .994571 / .405055 \\
\bottomrule
\end{tabular}
\end{table}

As a diagnostic, 670 chunks have stable arm outputs across numerical
configurations of the official reference (relative-$L_2$ threshold 0.1).
This selection uses no \flashrt{} outputs. On that subset, arm global/worst
cosine is .999691/.985367 for BF16 and .999188/.964353 for the mixed path.
The main table uses the unfiltered arm scores.

\subsubsection{Motus: four-input action and video agreement}
\label{app:motus-fidelity}
The Stage3 Motus deployment is checked on four RoboTwin2 input bundles using
the same ten-step FP4 configuration as the 167.08\,ms result, without TeaCache.
Each bundle supplies its first frame for calibration. Comparisons use the saved
upstream actions and decoded video; Table~\ref{tab:motus-four-inputs} reports
the per-input cosines, summarized over ten graph replays. The main fidelity
table reports the range across the four action cosines, rather than treating
replays of one input as additional samples.

\begin{table}[h]
\centering
\caption{Motus agreement on four RoboTwin2 inputs, compared with saved
upstream outputs. The evaluated configuration retains all ten denoising steps.}
\label{tab:motus-four-inputs}
\scriptsize
\setlength{\tabcolsep}{6pt}
\begin{tabular}{@{}lrr@{}}
\toprule
Input & Action cosine & Video cosine \\
\midrule
\code{sample\_00} & .999929 & .999117 \\
\code{sample\_01} & .999935 & .998644 \\
\code{sample\_02} & .999921 & .999144 \\
\code{sample\_03} & .999913 & .998844 \\
\bottomrule
\end{tabular}
\end{table}

\subsection{Deep Kernel Fusion Examples}
\label{app:fusion}

\paragraph{Calibrated producer--consumer quantization.}
An FP8 GEMM needs an FP8 input and a scale. In a dynamic path, separate kernels
may first find the activation maximum and then convert the activation before the
GEMM. For the $\pi_{0.5}$ encoder FFN, \flashrt{} measures an acceptable scale on
representative task inputs before capture. The preceding normalization kernel
then emits FP8 directly at that scale, and its single FFN consumer reads those
same bytes. The producer and consumer must be selected as a
pair: an FP8 output handed to a BF16 entry is an error, not a useful fallback.
Attention-output conversion is one boundary that still uses a separate kernel.

\paragraph{The ten-row action shape.}
The $\pi_{0.5}$ action expert always presents ten rows, but the stock NVFP4 GEMM
tile reserves space for 128 activation rows. Its out-of-bounds rows still cost
memory-system work. For selected decoder projections, \flashrt{} swaps the
GEMM operands: the large tile carries weights, the ten activation rows use
the short dimension, and a column-major
store has exactly the bytes the next kernel expects as row-major output. No
transpose kernel is needed. The fixed row count is guarded in the implementation.
The fixed deployment shape makes this tile choice valid throughout replay.

\paragraph{Phase fusion across layers.}
After an FFN down-projection, the policy must update a residual, normalize for
the next layer, and convert its input for the next GEMM. If those steps run as
separate kernels, each boundary adds a launch and moves an intermediate through
memory. \flashrt{}'s practical phase kernel combines the gated residual,
next-layer adaptive normalization, and FP8 conversion. The known layer and
denoising-step order tells it which next-layer style and scale to read.
Figure~\ref{fig:deep-fusion} counts 20 kernels and 193.2\,$\mu$s per layer on
the compiler path versus 12 kernels and 105.2\,$\mu$s on the composed
\flashrt{} path, measuring the combined pipeline changes at layer granularity.

\paragraph{QKV split, RoPE, and cache write.}
After the $\pi_{0.5}$ QKV projection, a conventional sequence can split the
combined output, rotate query and key positions, and write the key/value cache
in separate passes. \flashrt{} uses one kernel to do all three while those values
are already being visited. The pipeline fixes the head geometry and the K/V
cache layout; a device-side valid-prefix position tells the kernel where to
append K/V even when prompts are padded to a common capture shape. The cache
layout and position rule make the combined write valid throughout replay.

\paragraph{NVFP4 output directly from a GEMM.}
In the $\pi_{0.5}$ encoder feed-forward block, a GEMM produces two vectors,
\emph{gate} and \emph{up}. The earlier path launched a separate kernel to compute
$\operatorname{GELU}(gate)\mathbin{\times}up$, quantize the result, and pack it
for the next GEMM. \flashrt{} instead pairs gate and up weights when loading the
checkpoint. That puts each pair of results next to each other inside the GEMM,
allowing its epilogue to compute the product and write packed four-bit values
and block scales directly in the next GEMM's required layout.
An initial fused version wrote a full-width result
and doubled the next GEMM's weight traffic, so its feed-forward chain was slower
than the separate-kernel path. A compact, half-width store kept the next GEMM's
original shape; decision latency then fell, and the worst measured output cosine
improved because the intermediate was quantized once rather than twice.
The layout and producer--consumer format were designed
and validated for this pipeline, not inferred from an ordinary operator graph.

\subsection{Launch-Overlap Examples}
\label{app:overlap}

\paragraph{Fetching weights before the activation is ready.}
A GEMM cannot read an activation produced by the previous kernel until that
kernel finishes, but its fixed weights are already available. \flashrt{}'s
decoder GEMM starts streaming a few weight tiles before the wait, then loads
the activation only after the dependency is satisfied. This shortened the
decoder GEMM chain with identical outputs, but the gain was not monotonic: fetching all seven weight tiles early congested the memory
queue, while two or three tiles worked best. Moving the release signal from
the load warp to the multiply warp also gave up much of the benefit. The
selected variant streams only the profitable prefix.

\paragraph{Choosing between persistent and separate GEMMs.}
\flashrt{} also built a resident grid that runs several decoder GEMMs in one
launch. Its load warp fetches the next GEMM's weights while the current one
finishes. With independent GEMMs, the chain ran faster than separate launches
and came close to the measured memory-read ceiling. The version that also
performs the required normalization between GEMMs, however, was slower than
separate launches. The next GEMM must wait for the normalization result, and a
grid-wide rendezvous adds cost that weight prefetch cannot hide. The
implementation passed bit-exact checks but is disabled in the shipped path.
A faster isolated chain was not a faster pipeline here; the useful rule is to
fuse the memory-bound work \emph{between} GEMMs, not every GEMM into one kernel.

\paragraph{Knowing when overlap has reached its limit.}
More early work is not always better. In Thor tuning, extending early launch
to RoPE, softmax, and FP8-conversion kernels produced no further measurable
change. Prefetching the next layer's weights into cache also slowed every
tested variant of the decoder chain, because the prefetch competed with demand
traffic. Together with the rejected
persistent sequence above, these cases bound the technique: it can hide
launch setup or input-independent memory traffic, but not work whose result
the successor actually needs. \flashrt{} therefore tests overlap in the full
pipeline as well as in isolated chains.

\subsection{Execution-State Reuse Example}

\paragraph{Instruction state.}
The $\pi_{0.5}$ action expert has ten denoising steps and eighteen layers; its
adaptive-normalization values form a finite step-by-layer table for a given
instruction. Rebuilding those values inside every decision repeats work whose
input has not changed. \flashrt{}'s \code{set\_prompt} builds the table and prompt
embedding once, keeps them in named buffers, and lets later kernels select the
needed row. A same-size prompt update overwrites the existing embedding rather
than replacing its address; a device-side valid length masks padding without
recapture. In an earlier compiler implementation, in-graph Dense GEMMs rebuilt
these styles at every inference, and hand-editing that graph to precompute them
reduced its latency. \flashrt{} uses the same temporal-hoisting idea in its prompt
setup. The instruction lifetime and refresh rule specify when this prepared
state can be reused and when it must be rebuilt.

\section{Limitations and Future Work}
\label{app:limitations}
\flashrt{} targets small-batch, fixed-shape physical-AI inference, and several
limitations follow directly from this scope.

\textbf{Integration cost.} New shapes require recapture, and a new native
integration requires model--hardware pipeline co-design, tuning, calibration, and
validation (Appendix~\ref{app:examples}); this is engineering work, not automatic
discovery. Shared kernels and contract interfaces reduce repeated implementation,
but each new model--hardware pipeline is still designed and validated per model.

\textbf{Evaluation scope.} The current evaluation reports warmed-up inference on
two Blackwell GPUs, with four policies on Thor and six on RTX 5090, and task
success for $\pi_{0.5}$ and LingBot-VA in LIBERO simulation. Results use fixed or
recorded inputs offline; physical-robot behavior and complete sensor-to-actuator
deadlines are not measured here.

\textbf{State and capsules.} Capsules resume compatible sessions within a single
process (\S\ref{sec:capsule}); captured graphs contain device addresses, so
cross-process or cross-device resumption is out of scope.

\textbf{Directions for further evaluation.} Broader hardware and task coverage,
sustained control-loop characterization at the real observation/action-chunk
cadence, and multi-session resource accounting remain open.

\end{document}